%% file: main.tex
\documentclass[11pt]{article}
\usepackage[preprint]{acl}
\usepackage{times}
\usepackage{latexsym}
\usepackage[T1]{fontenc}
\usepackage[utf8]{inputenc}
\usepackage{microtype}
\usepackage{graphicx}
\usepackage{booktabs}
\usepackage{amsmath,amssymb}
\makeatletter
\newcommand{\tablecaption}{\def\@captype{table}\captionsetup{hypcap=false}\caption}
\newcommand{\widepage}[1]{\clearpage\global\@colroom\@colht\global\vsize\@colht\twocolumn[{#1}]}
\makeatother
\title{One Model, Two Paths: Balancing Single-Pass and Looped Performance in Language Models}
\author{
  Jiarun Shen$^{1}$ \quad Lei Zhao$^{2}$ \quad Xiao Li$^{1}$ \\
  $^{1}$The Chinese University of Hong Kong, Shenzhen \\
  $^{2}$Shanghai Jiao Tong University \\
  {\small\texttt{jiarunshen@link.cuhk.edu.cn}} \\
  {\small\texttt{l.zhao@sjtu.edu.cn} \quad \texttt{lixiao@cuhk.edu.cn}}
}
\begin{document}
\maketitle
\raggedbottom
\begin{abstract}
Repeating Transformer layers can improve language-model performance without adding parameters.
A pretrained model can be adapted to support both single-pass and looped inference by reusing middle-layer weights for additional computation.
Because both paths share weights, training to improve looped predictions can degrade single-pass quality.
We study joint training strategies using weighted losses on both outputs, with guidance from the frozen original model.
Joint supervision substantially reduces single-pass degradation compared to looped-only training, while preserving looped gains over ordinary continued training.
Most of this benefit comes from adjusting the single-pass loss weight, with guidance from the frozen original model further improving the looped output.
With validation-selected weights, our method preserves single-pass quality within a predefined tolerance of ordinary continued training while improving looped predictions.
\end{abstract}

\section{Introduction}
Repeating Transformer layers gives a language model additional computation without requiring another set of layer weights. Recent work has shown that pretrained models can be adapted to benefit from such repeated computation \citep{koishekenov2025etd,mcleish2025retrofitted,marchenko2026closing}. The adapted model retains its single-pass path, which applies each layer once, alongside a looped path that reuses the same weights to repeat a subset of layers. The same model can then use the single-pass path when computation is limited and the looped path when additional computation is available.

We study adaptation that preserves all pretrained layers and the original output head. Because both paths share all weights and remain trainable, updating the model to improve looped predictions also changes the weights used by the single-pass path. The training challenge is to preserve single-pass quality while making the looped path useful.

Evaluating whether adaptation succeeds requires comparing against two baselines: the starting checkpoint and the alternative of continuing to train the original model without looping. We call the latter ordinary continued training. Keeping the starting quality is not enough if the same budget would improve the original model further.

A natural training approach is to supervise both outputs with next-token labels. We also consider adding a teacher signal: using one prediction's probability distribution to guide the other prediction during training. This resembles distillation in standard teacher-student setups, adapted here to two paths that share all weights. We compare weighted label losses on both outputs with and without this teacher signal, starting either from the initial checkpoint or after an initial period of training only the looped path. The staged setting allows us to examine how single-pass quality can be recovered while preserving useful looped predictions.

Adding a teacher signal introduces a different supervision target for one output. An improvement over training both paths with labels alone might also be achievable by giving that output's label loss more weight. We include a stronger label-only baseline to test this possibility, with its weight chosen to account for the added teacher term (\S\ref{sec:matched-control}). This comparison measures how much of the observed benefit the teacher signal provides beyond stronger label supervision.

Training cost also matters, because computing the teacher signal takes time even when two methods execute the same number of model layers. We therefore compare methods under separate budgets based on layer evaluations and measured training time, including any teacher computation.

The teacher signal can come from different sources. One option uses the current looped output as teacher for the single-pass output, recomputing its distribution at each training step. Another option uses a frozen copy of the original model, which provides a fixed reference throughout adaptation. We compare these two sources and examine whether the teacher signal primarily benefits the output that receives it directly or also improves the other output through their shared weights.

We make three main contributions:

\begin{itemize}
\item \textbf{Joint supervision reduces single-pass degradation and supports recovery.} We find that applying a small loss weight to the single-pass output substantially reduces the degradation caused by training only the looped path. During recovery, supervising both outputs preserves more looped quality than returning to single-pass training alone, although it recovers less single-pass quality.

\item \textbf{Increasing the single-pass loss weight reproduces most of the benefit from adding a teacher signal to guide the single-pass output.} Without a teacher signal, adjusting the loss weight on the single-pass output alone recovers most of the observed improvement. After validation-based weight selection, adding the teacher signal shows no additional mean improvement over the selected label-only baseline at matched training time.

\item \textbf{A frozen teacher from the original model can preserve single-pass quality while improving the looped output.} We call this method Frozen-original KD. With a validation-selected label weight, it uses the original model's predictions to guide the single-pass output. In the main setting, this method preserves single-pass quality within a predefined tolerance of ordinary continued training while improving looped predictions over that baseline.
\end{itemize}

The results separate two training choices: the loss weight on each output and the source of the teacher signal. With validation-selected label weights, Frozen-original KD preserves single-pass quality within the predefined tolerance of ordinary continued training while improving looped predictions over that baseline in the main setting. This joint result does not extend to the secondary WikiText corpus. Evaluating both outputs against ordinary continued training is necessary to distinguish a useful supervision choice from successful retention of both prediction paths.

\section{Related Work}
\paragraph{Recurrent conversion.} Encode, Think, Decode repeats pretrained middle layers while retaining the physical stack \citep{koishekenov2025etd}. Retrofitted Recurrence studies conversion and training curricula under compute constraints \citep{mcleish2025retrofitted}, while Closing the Loop uses input mixing and smoothed exit-loss weights \citep{marchenko2026closing}. \citet{shapiro2026retrofitting} studies a recurrent retrofit with an identity-preserving one-loop path. These works adapt pretrained models to benefit from looped computation, establishing the feasibility of converting existing models without retraining from scratch.

\paragraph{Supervision across depths.} RRT combines intermediate label losses with detached deeper-output distillation in compressed recursive models \citep{bae2025relaxed}. LoopFormer uses trajectory conditioning and stop-gradient consistency across computation budgets \citep{jeddi2026loopformer}. Concurrent work studies frozen and dynamic cross-loop teachers through on-policy distillation for mathematical reasoning, using student-generated rollouts and reverse KL \citep{wang2026recurrent}. These works supervise models at multiple computation depths. Our setting differs in that we retain the complete single-pass path alongside the looped path, with both paths sharing all weights, and use fixed-text next-token supervision with forward KL.

\paragraph{Distillation and retention controls.} Guidance from an existing predictor and soft-target regularization have prior precedents \citep{li2016learning,yuan2020revisiting}. Distillation and label smoothing need not have identical effects \citep{sultan2023distillation}. We compare single-pass label strength and teacher source for two paths that share all weights. Both the single-pass and looped outputs are evaluated against ordinary continued training, with teacher computation included. The coefficient-matched label control tests how much of the improvement from adding a teacher signal can be reproduced by stronger label supervision alone. It does not identify the remaining difference with a particular mechanism.

\section{Training Strategies}
\label{sec:methods}
\subsection{Two Predictions from Shared Weights}
The decoder has eight Llama-style blocks \citep{touvron2023llama2}. Let $E$ contain the embedding and first two blocks, $M$ the middle four blocks, $D$ the last two blocks, and $H$ the final normalization and vocabulary head. The prediction paths are
\begin{align}
f_s(x;\theta)&=H(D(M(E(x)))),\notag\\
f_c(x;\theta)&=H(D(M(M(E(x))))).
\label{eq:paths}
\end{align}
The single pass applies all eight blocks once. The looped path applies twelve blocks. At the starting weights $\theta_0$, $f_s$ is exactly the original model. Both paths share all parameters, and all weights remain trainable. Joint computation shares the first six block applications before branching, so one joint update uses fourteen forward-block applications. Appendix~\ref{app:repro} gives the architecture and implementation.

\subsection{Labels, Distillation, and a Matched Control}
\label{sec:matched-control}
Let $\mathcal L_s$ and $\mathcal L_c$ be the single-pass and looped cross-entropies, averaged over valid targets. The Joint CE baseline and Online KD objective are
\begin{align}
\mathcal L_{\rm CE}&=0.1\mathcal L_s+0.9\mathcal L_c,\label{eq:ce}\\
\mathcal L_{\rm KD}&=\mathcal L_{\rm CE}+0.1\operatorname{KL}(\operatorname{sg}[p_c]\|p_s).
\label{eq:kd}
\end{align}
Online KD uses the current model's looped output as its teacher. Its distribution is recomputed at each update and detached in the KL term. The looped label loss remains fully differentiable. Both distributions use temperature 1, and KL covers the full vocabulary, averaged over valid targets. This uses the current looped output to guide the single-pass output, distinguishing it from guidance by a frozen predictor at the initial checkpoint.

Our coefficient-matched label control, Matched CE, uses
\begin{equation}
\mathcal L_{\rm matched}=0.2\mathcal L_s+0.9\mathcal L_c.
\label{eq:matched}
\end{equation}
The coefficients are not normalized by 1.1. For one target with label vector $e_y$, the single-pass logit gradients of KD and matched CE are $0.2p_s-0.1e_y-0.1p_c$ and $0.2p_s-0.2e_y$. Matching the coefficient on $p_s$ controls one change introduced by KL. It does not match gradient direction, norm, clipping, or parameter updates. This tested coefficient-matched control is one label-weight setting, not an optimized label-supervision frontier. Appendix~\ref{app:diagnostics} examines those differences.

Ordinary and loop-only training optimize $\mathcal L_s$ and $\mathcal L_c$ separately. A lightly weighted joint baseline uses $0.05\mathcal L_s+0.95\mathcal L_c$. Joint CE, KD, and matched CE start either at the ordinary parent or after 2,048 loop-only updates. The staged ordinary control instead follows the same switch state with ordinary-only updates. The ten methods and their update counts appear in Table~\ref{tab:main}.

\subsection{Our Method: Frozen-original KD}
\label{sec:targets}
Our method, Frozen-original KD, guides the current single pass using a frozen copy of the original single-pass predictor. We retain next-token label losses on both current outputs and use
\begin{equation}
\mathcal L(q;\alpha)=\alpha\mathcal L_s+0.9\mathcal L_c+0.1\operatorname{KL}(\operatorname{sg}[q]\|p_s).
\label{eq:target}
\end{equation}
For our method, $q=p_s(\theta_0)$. The frozen teacher performs an independent forward pass; all student weights remain shared and trainable. Teacher computation is included in the training budgets, and the teacher is not used at inference. Teacher-source controls replace $q$ with the current looped prediction $p_c(\theta_t)$ or the frozen initial looped prediction $p_c(\theta_0)$, keeping both label weights and the KL coefficient unchanged. The initial teacher-source study uses $\alpha=0.1$. In the label-weight study, validation selects $\alpha=0.3$ for Frozen-original KD; the teacher and KL coefficient remain unchanged.

\textbf{Additional controls.} Target-content controls flatten or permute non-label probabilities. Both preserve the gold-token probability. Permutation also preserves entropy and the probability multiset, changing their assignment to vocabulary tokens. Early-only guidance uses online KD for 512 updates, then matched CE; delayed guidance reverses this order. Both label objectives remain active throughout. Appendix~\ref{app:targets} gives all seven added conditions and the fixed permutations.

For label-weight selection, we compare
\begin{align}
\mathcal L_{{\rm CE},m}&=m\mathcal L_s+0.9\mathcal L_c,\notag\\
\mathcal L_{{\rm KD},m}(q)&=(m-0.1)\mathcal L_s+0.9\mathcal L_c\notag\\
&\quad+0.1\operatorname{KL}(\operatorname{sg}[q]\|p_s),
\label{eq:grid}
\end{align}
with $m\in\{0.2,0.4,0.8,1.6\}$. Here $m$ is the nominal coefficient on the single-pass distribution in its logit gradient, not always its label-loss coefficient. Coefficients are not normalized. Equal $m$ does not imply equal full gradients, clipping, or parameter updates.

\section{Experimental Setup}
\label{sec:setup}
\subsection{Models and Data}
The main experiments adapt two fixed 133,830,912-parameter models, C and D, after 32,768 ordinary updates. In the initial supervision and target studies, each model has three training-data orders, giving six paired units per method. Within a unit, methods restore the same weights, optimizer buffers, random state, and data position. Orders permute the same windows within predetermined segments. The window sets agree at every registered segment boundary. The six units are repetitions of adaptation within two parents, not six independent pretraining runs. Historical continuations are excluded from these summaries.

Training uses a two-billion-token pool of English FineWeb-Edu text \citep{penedo2024fineweb}, a GPT-2 tokenizer, sequence length 1,024, and batch size 32. A fresh validation set contains 524,288 targets in sixteen blocks. A separate confirmation set contains 2,097,152 targets in thirty-two blocks. Exact document hashes exclude recorded training and earlier evaluation text. Previously inspected WikiText \citep{merity2017pointer} supplies 282,624 targets in eighteen clusters as a secondary corpus. The supervision study uses this confirmation set for all primary NLL comparisons. Appendix~\ref{app:assets} reports counts and data terms.

A separate target study adds seven conditions in the same six parent--order units, giving 42 new endpoints. Fresh validation and confirmation text have the same target counts as above but different document sets. Existing controls are re-evaluated on the new text. These interventions reuse the two parents and six orders; they add no pretraining seeds. The two studies retain separate result tables. The subsequent label-weight study uses new validation and confirmation documents with the same target counts; Appendix~\ref{app:weight-study} distinguishes the three studies.

\subsection{Training Methods and Schedules}
\label{sec:training-methods}
Names prefixed by \textbf{Loop$\to$} apply the named objective after 2,048 loop-only updates. Thus Loop$\to$KD uses joint labels and online distillation during recovery. Figures call the schedules \emph{From start} and \emph{After loop-only}; appendix tables call them static and staged. The separate 512-update guidance controls retain both label objectives from the start.

The label-weight study compares the three strategies over the same four-point grid. Selection uses validation results from C1 and D1 at the fixed training-time endpoint. Within each family, both single-pass gaps must meet their retention margins on each development order. We then select the admissible weight with the lowest mean looped validation NLL and evaluate looped utility after selection. The chosen weight is fixed before evaluation on C2, C3, D2, and D3, using new confirmation text and paired fixed-time ordinary endpoints. Appendix~\ref{app:weight-study} gives the thresholds, tie-breaking rule, and handling of infeasible families.

\subsection{Training and Evaluation Budgets}
Muon and auxiliary AdamW retain the inherited learning rates 0.004 and 0.00006 throughout adaptation and stage switches \citep{jordan2024muon,loshchilov2019adamw}. The initial studies use six RTX 5090 GPUs with 32\,GB each; the label-weight study and additional configurations use eight. Each training process uses one GPU.

The forward-block cap is 40,960 applications at fixed batch size. Ordinary, loop-only, and joint updates use 8, 12, and 14 applications. A staged joint path uses $2,048\times12+1,170\times14=40,956$ applications. An ordinary suffix uses $2,048\times12+2,048\times8=40,960$. Different schedules therefore receive different update counts and training exposure.

The second cap is 2,227.587 seconds of synchronized forward computation, loss, backward computation, gradient clipping, and optimizer work. Loading, evaluation, and checkpoint I/O are excluded. Staged time includes the newly trained loop-only prefix. Every trajectory covers the cap, and its time endpoint is the last saved update within the cap. The two endpoints come from the same training trajectory. GPU assignments rotate across paired units and method rounds. Appendix~\ref{app:resources} separates method timing from total experiment costs.

Target interventions are first compared at 2,925 student updates. Equal updates fix student exposure but do not equate teacher cost. Frozen looped and single-pass teachers add 12 and 8 block applications per update, giving totals of 26 and 22. At the 40,960-block cap, their endpoints are 1,575 and 1,861 updates. Time comparisons include teacher computation and reused prefixes.

\begin{figure*}[t]\centering
\includegraphics[width=\textwidth]{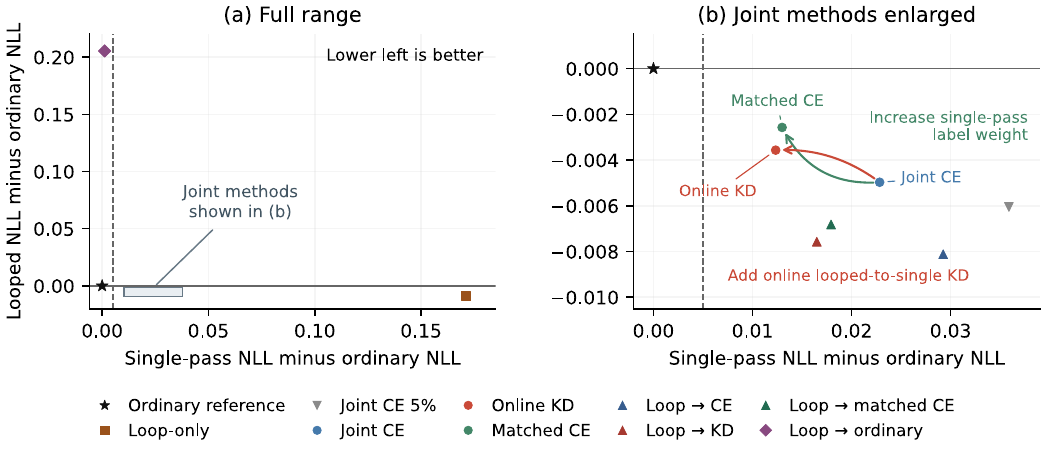}
\caption{Prediction quality relative to ordinary continued training on the supervision-study confirmation text at the forward-block cap. Axes show single-pass and looped NLL minus ordinary single-pass NLL, in nats per target. Lower left is better. Joint CE uses labels on both paths, Matched CE increases the single-pass label weight, and Online KD adds a teacher signal using the current looped output. Panel (b) enlarges the joint methods boxed in panel (a). Points average six paired continuations. Arrows compare endpoints; they are not training trajectories or a causal decomposition. The star marks the reference origin, not a measured ordinary looped output. The horizontal line marks equal looped and ordinary NLL; the vertical dashed line marks the 0.005 single-pass tolerance. Other methods and Loop$\to$ schedules are defined in \S\ref{sec:matched-control} and \S\ref{sec:training-methods}. Frozen-original KD (ours) is evaluated in Figure~\ref{fig:increment}.}
\label{fig:tradeoff}
\end{figure*}

\subsection{Retention and Uncertainty}
\paragraph{Evaluation metric.} We report average next-token negative log-likelihood (NLL), $\ell(p)=-N^{-1}\sum_{i=1}^{N}\log p(y_i\mid c_i)$, where $y_i$ is a reference token, $c_i$ is its available reference context, and $N$ is the number of valid evaluated targets. This is the standard hard-label cross-entropy loss, using natural logarithms and measured in nats per target token; lower is better. Each path is evaluated separately, without label-loss weights or the distillation term.

Let $\ell_s$ and $\ell_c$ denote single-pass and looped NLL. For adapted weights $\theta$ and an ordinary continuation $\theta_{s,b}$ under budget $b$, define
\begin{align}
x&=\ell_s(\theta)-\ell_s(\theta_{s,b}),\notag\\
y&=\ell_c(\theta)-\ell_s(\theta_{s,b}).
\label{eq:comparisons}
\end{align}
Negative differences favor the adapted model. Starting retention is $D_{\rm start}=\ell_s(\theta)-\ell_s(\theta_0)$. The within-model looped difference is $d=\ell_c(\theta)-\ell_s(\theta)=y-x$. The fixed requirements are $D_{\rm start}\leq0.005$, $x\leq0.005$, $y<0$, and $d<0$. Together they test retention of the original predictor, ordinary continued training gains, and useful additional looped computation.

The margins are predeclared reporting scales, not deployment guarantees. An NLL increase of 0.005 corresponds to approximately 0.50\% higher perplexity. The target study uses the narrower $\pm0.001$ scale, about 0.10\%, for interval-based practical equivalence. Neither threshold establishes downstream capability preservation. Appendix~\ref{app:sensitivity} gives algebraic conversions and a supplementary sensitivity analysis of the existing intervals.

Means in the initial two studies weight the six paired units equally, with half the weight on each parent. Paired, token-weighted block bootstrap intervals use 20,000 shared resamples \citep{dror2018hitchhiker}. Direct comparisons have four prespecified families of two or four effects, giving 97.5\% or 98.75\% adjusted marginal intervals. These intervals describe text uncertainty conditional on the fitted models. Separate within-parent data-order summaries report all six differences, means, standard deviations, and small-sample intervals conditional on C/D. The two sources of uncertainty are not pooled. Appendix~\ref{app:stats} gives the full comparisons and statistical definitions. The target study retains separate content, source, and timing families of 6, 8, and 10 effects, with adjusted text intervals of 99.1667\%, 99.375\%, and 99.5\%. Appendix~\ref{app:targets} reports these intervals alongside conditional data-order intervals.

Selected-recipe means use the four held-out continuation orders. Full-grid summaries also include the two development orders and are descriptive. For the three selected recipes and four criteria, the prespecified twelve-effect correction gives 99.5833\% text intervals. Fixed-parent data-order 95\% intervals remain separate (Appendix~\ref{app:weight-study}).

\begin{figure*}[t]\centering
\includegraphics[width=\textwidth]{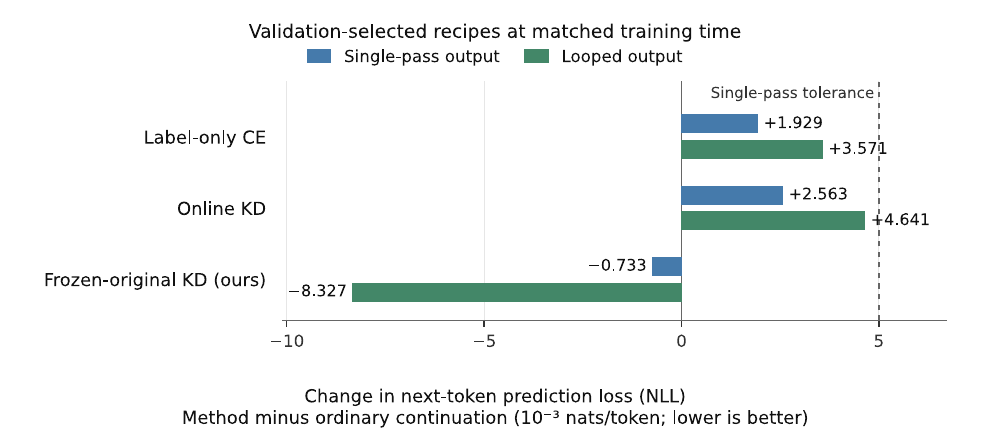}
\caption{Both outputs after validation-based label-weight selection. Bars show four-order mean NLL differences from fixed-time ordinary continuation on the new confirmation text, including teacher computation in the budget. Within each method, the upper bar is single-pass and the lower bar is looped. Lower is better. The dashed line marks the single-pass retention tolerance; it is not a looped-gain threshold. Bars are descriptive means; Table~\ref{tab:target-primary} reports the retention criteria and key uncertainty intervals, with full results in Appendix~\ref{app:weight-study}.}
\label{fig:increment}
\end{figure*}

\section{Results}
\subsection{Joint Supervision Retains More Single-Pass Quality}
\label{sec:results}
Table~\ref{tab:main} and Figure~\ref{fig:tradeoff} show all ten methods at the forward-block cap. Loop-only training improves looped NLL over ordinary continued training by 0.009048, but worsens single-pass NLL by 0.171239. Giving the single-pass label loss 5\% weight reduces that gap to 0.035882 while retaining a looped improvement of 0.006052. Both loop-only and lightly weighted joint training improve the looped prediction over ordinary continued training in all six paired units. Their adjusted text intervals exclude zero.

\begin{table}[t]\centering\small
\setlength{\tabcolsep}{3pt}
\input{tables/presentation_v2/r43_endpoints_confirm_proxy.tex}
\caption{Confirmation NLL at the forward-block cap, averaged over three new data orders for each of two fixed parents. Lower is better. Staged update counts include the 2,048-update loop-only prefix. The ordinary reference uses its single pass; the unused ordinary looped output is omitted.}
\label{tab:main}
\end{table}

Keeping the looped loss during recovery also changes the trade-off. At the same forward-block cap, the joint CE suffix lowers looped NLL by 0.213603 relative to the ordinary suffix and raises single-pass NLL by 0.028091. The directions agree across all six paired units. The ordinary suffix restores more single-pass quality, while joint recovery retains more looped quality.

\subsection{Stronger Labels Reproduce Most of the KD Increment}
\label{sec:matched}
At the forward-block cap, Online KD lowers single-pass NLL relative to Joint CE by 0.010496 from the start and 0.012754 after loop-only training. Matched CE reproduces most of this increment (0.009847 and 0.011321). Because the objectives also change clipping and optimization, these ratios are not causal attributions.

KD retains smaller benefits over Matched CE, with all four adjusted text intervals below zero. Static KD improves both outputs in all six orders; staged KD improves the looped output in five, so a favorable mean need not hold in every continuation. Appendix~\ref{app:stats} retains every direct effect and order, and Appendix~\ref{app:diagnostics} reports output and local update diagnostics.

\subsection{The Residual Depends on the Resource Budget}
\label{sec:time}
In the isolated benchmark, KD takes 20.9\% more update time than Joint CE and 15.3\% more than Matched CE, despite equal forward-block counts. A balanced follow-up removes the CE--Matched CE timing gap under a common weighted-CE implementation but reproduces it with the historical implementations; the difference is implementation-dependent (Appendix~\ref{app:resources}). Comparisons retain the actual trajectories' complete-update times.

At the time cap, KD's mean single-pass differences from Matched CE are $-0.001025$ from the start and $-0.001009$ after loop-only training. Adjusted text intervals are below zero, while conditional order intervals span zero. Looped differences are $-0.000006$ and $-0.000456$, with text intervals spanning zero. The residual is unresolved across continuation orders at this budget; the results do not establish equivalence. Time-matched static KD instead has higher single-pass NLL than Matched CE on WikiText. Appendix~\ref{app:stats} reports both corpora, all effects, and separate uncertainty intervals.

\begin{table*}[t]\centering\small
\setlength{\tabcolsep}{4pt}
\input{tables/complete_report_20261011/selected.tex}\par\smallskip
\input{tables/complete_report_20261011/key_intervals.tex}
\caption{Validation-selected recipes at the complete-update time cap, including teacher computation. All values are in $10^{-3}$ nats per target on the new confirmation text. The upper panel gives four-order means for the criteria in Eq.~\ref{eq:comparisons}: $D_{\rm start}\leq5$, $x\leq5$, $y<0$, and $d<0$ in these units. All four counts point-estimate successes, not simultaneous per-order guarantees. The lower panel highlights paired Frozen-original KD contrasts. The first two use the twelve-effect text family (99.5833\%); the third uses the four-effect direct-comparison family (98.75\%). Order intervals are separate conditional 95\% intervals. Full results appear in Appendix~\ref{app:weight-study}.}
\label{tab:target-primary}
\end{table*}

\subsection{A Frozen Original Teacher Primarily Improves the Looped Output}
\label{sec:target-results}
At the initial fixed coefficients ($\alpha=0.1$), Frozen-original KD primarily improves the looped output: NLL falls by 0.009884 relative to Matched CE at matched training time, including teacher computation. Its single-pass difference is $+0.000537$. The frozen initial looped teacher instead harms the single pass while improving the looped output. Teacher source changes both predictions, including the output without a direct KL term.

Appendix~\ref{app:targets} retains the fixed-coefficient comparisons, target-content and timing controls. The fixed Frozen-original KD recipe still has a 0.012772 single-pass gap from validation-selected ordinary continuation (Table~\ref{tab:target-retention}), a different reference from Matched CE.

\subsection{Label-Weight Selection and Joint Retention}
\label{sec:selection}
The initial fixed-coefficient studies do not identify a configuration meeting all four criteria on their primary confirmation texts (Appendices~\ref{app:selection} and~\ref{app:targets}). We now compare the recipes selected under the common rule in \S\ref{sec:setup}. Label-only CE and Online KD select $m=0.8$, while Frozen-original KD selects $m=0.4$.

On the four held-out continuation orders, selected Frozen-original KD meets all four point-estimate criteria in each order. Family-adjusted text intervals for the four-order mean also support all four criteria (Table~\ref{tab:target-primary}; Appendix~\ref{app:weight-study}). The single-pass interval relative to ordinary continuation crosses zero, supporting retention within the predefined margin rather than superiority. Unlike selected label-only CE, the frozen-teacher recipe retains useful looped predictions (Figure~\ref{fig:increment}).

Relative to selected CE, its looped NLL improves by 0.0119, with a conditional data-order interval of $[-0.0202,-0.0036]$. The separately adjusted text interval supports the same direction. Selected Online KD shows no additional mean improvement over selected CE. These comparisons concern the tested weight grid, not a globally optimal label-only objective.

\subsection{Receiving Path and Additional Configurations}
\label{sec:extensions}
With the same frozen teacher and total KL weight, guidance to the single pass, looped output, or both improves mean looped NLL relative to its respective label control. Direct looped guidance also improves the single pass relative to its own control. The effect is therefore not unique to single-pass guidance. The controls use the original low single-pass label weight and do not identify the best receiving path for the selected recipe.

At the original fixed coefficients, the frozen teacher also improves looped NLL over the label control in a deeper $8\to16$ configuration and native Parcae. Neither extension meets the full joint criteria. These are fixed-recipe extensions, not tests of the selected label weight from the main setting. Appendix~\ref{app:weight-study} reports their separate configurations and results.

\subsection{Task Accuracy and Retained Correct Predictions}
\label{sec:tasks}
We evaluate both outputs on ARC-Easy, PIQA, and HellaSwag \citep{clark2018arc,bisk2020piqa,zellers2019hellaswag}. In the initial supervision study, joint supervision with Online KD after loop-only training improves ARC-Easy accuracy, but task directions are mixed. Increased accuracy can also accompany more errors on examples the original predictor answered correctly. Appendix~\ref{app:tasks} reports that study's correctness transitions and separate item and training-order intervals.

For the selected recipes, Frozen-original KD improves mean ARC-Easy accuracy over ordinary continuation but has no consistent advantage across tasks or over selected CE. On the original predictor's correct ARC-Easy examples, it reduces the loss of correct predictions relative to ordinary continuation by 3.53 percentage points for the single pass and 2.83 for the looped output. Paired item intervals support both reductions, whereas conditional training-order intervals span zero. HellaSwag accuracy remains lower on average than ordinary continuation. These are task-specific retention gains for the fitted models, not stable improvement across orders or tasks (Appendix~\ref{app:weight-study}).

\section{Analysis}
\paragraph{Label strength and teacher source.} Increasing the single-pass label weight narrows the gap from ordinary continuation but can remove the looped advantage. The full grid shows that Frozen-original KD balances the two outputs at an intermediate weight (Appendix~\ref{app:weight-study}). Guidance from the original model follows the use of old-model outputs during continued learning \citep{li2016learning}; its value here depends on its combination with label supervision. The receiving-path controls show that the looped benefit is not unique to guiding the single pass. They do not rank the receiving paths for the selected recipe because their label controls differ and their weights were not jointly selected.

\paragraph{Additional computation is not uniformly beneficial.} Extra latent iterations can turn correct predictions into errors \citep{fu2025think}; our diagnostic also contains both corrections and new errors (Appendix~\ref{app:diagnostics}). The selected recipe meets the primary text criteria but both outputs have higher NLL than ordinary continuation on WikiText. This does not isolate repeated computation as the cause or identify a noise-amplification mechanism. The gain over selected label-only training is measured at the same looped inference depth. Whether using the looped rather than single-pass output justifies its extra inference cost is a separate deployment question, not established by training-time comparisons.

\paragraph{Retention depends on the reference.} Ordinary continuation measures adaptation's opportunity cost. New fixed-time and initial validation-selected references involve different recipes, texts, and averaging units; subtracting their gaps is not a controlled improvement. Retention within 0.005 is distinct from equal NLL or superiority. Updates can change which examples are correct even when accuracy improves \citep{xie2021regression}, motivating separate loss and correctness-retention measures.

\section{Conclusion}
\textls[-10]{Adapting a pretrained model for repeated computation requires evaluating both its original and looped predictions. Stronger single-pass labels reproduce most of the improvement from adding a teacher signal in the initial comparisons. With validation-selected weights, Frozen-original KD retains single-pass NLL within the predefined margin of ordinary continued training and improves looped NLL over that reference in the main setting. Its looped output also improves over selected label-only training at matched training time. The results support comparing label strength and teacher source jointly. The corpus and task results delimit this conclusion to the evaluated text-retention setting.}

\clearpage
\section*{Limitations}
The main experiments study two fixed 133.8M-parameter pretrained models on English text, with additional tests of recurrence depth and a native recurrent architecture. The comparisons use a finite set of label weights and teacher configurations under specified training budgets, with uncertainty reported separately for evaluation data and continuation orders. Experiments at larger model scales, in other languages, and under generation-time budgets are natural directions for further study.

\section*{Ethical Considerations}
The study uses existing English corpora and public benchmarks without new human participants or annotation. Web text can contain biased, harmful, copyrighted, or personally identifying material. Likelihood evaluation and exact deduplication do not remove those risks. We conducted no separate personal-information or harmful-content audit. Appendix~\ref{app:assets} describes data and software terms. Appendix~\ref{app:resources} reports the available compute records. Energy use and emissions were not measured.

\newpage
\begin{samepage}
\section*{AI Statement}
Generative AI tools were used primarily for language polishing and grammar checking. They also assisted with checking manuscript formatting. The authors retain full responsibility for the scientific content and the final manuscript.
\end{samepage}
\clearpage
\bibliography{references}
\clearpage
\appendix

\widepage{
\section{Reproducibility Details}
\label{app:repro}
\begin{minipage}{\textwidth}\centering\small
\begin{tabular}{llll}
\toprule
Setting & Value & Setting & Value \\
\midrule
Parameters & 133,830,912 & Physical blocks & 8 \\
Hidden / MLP size & 768 / 2,048 & Attention / KV heads & 12 / 12 \\
Vocabulary & 50,257 & Repeated block indices & 2--5 (zero-based) \\
Sequence / batch & 1,024 / 32 & Microbatch / accumulation & 4 / 8 \\
Arithmetic / weights & BF16 / FP32 & Gradient norm limit & 1 \\
Optimizer & Muon NS5 + AdamW & Muon / AdamW learning rates & 0.004 / 0.00006 \\
Muon momentum & 0.95 & AdamW $\beta$, $\epsilon$ & $(0.9,0.95)$, $10^{-8}$ \\
Weight decay & 0.1 & Zero-decay parameters & One-dimensional \\
Parent updates & 32,768 & Parent initialization seeds & 2901, 3601 \\
C data-order seeds & 43001--43003 & D data-order seeds & 43004--43006 \\
GPU type & RTX 5090 32\,GB & GPUs / process & 6 total / 1 per process \\
Forward-block cap & 40,960 & Complete-update time cap & 2,227.587 seconds \\
\bottomrule
\end{tabular}
\tablecaption{Main-study model and training configuration. Method-specific continuation counts are in Table~\ref{tab:main}. Learning rates and optimizer states are inherited and retained through stage switches. Appendix~\ref{app:resources} reports GPU hours. Related architectures use depth sharing \citep{dehghani2019universal,lan2020albert}, block recurrence \citep{hutchins2022block}, recurrent depth \citep{geiping2025recurrent}, or layer dropping \citep{fan2020layerdrop}.}
\label{tab:config}
\end{minipage}\par\medskip
\begin{minipage}{\textwidth}\centering\small
\begin{tabular}{lrrrl}
\toprule
Dataset / split & Documents or items & Targets & Blocks & Role \\
\midrule
FineWeb-Edu training & 1,935,463 & 2,000,000,435 tokens & -- & Shared training pool \\
Supervision validation & 551 & 524,288 & 16 & Checkpoint selection \\
Supervision confirmation & 2,190 & 2,097,152 & 32 & Primary text evaluation \\
WikiText test & 62 articles & 282,624 & 18 & Secondary text evaluation \\
ARC-Easy test & 2,376 & -- & -- & Task evaluation \\
PIQA validation & 1,838 & -- & -- & Task evaluation \\
HellaSwag validation & 10,042 & -- & -- & Task evaluation \\
\bottomrule
\end{tabular}
\tablecaption{Supervision-study data and evaluated task splits. Target-study documents are listed separately in Appendix~\ref{app:targets}. Public task validation splits serve as final evaluations and do not select checkpoints.}
\label{tab:task-data}
\end{minipage}\par\medskip
}

\subsection{Models, Data, and Licenses}
\label{app:assets}

The decoder uses RMSNorm ($\epsilon=10^{-5}$), rotary base 10,000, SiLU, untied embeddings, no attention/MLP biases or dropout, and 4,096 maximum positions. No released Llama~2 weights are used. The \href{https://huggingface.co/openai-community/gpt2/blob/607a30d783dfa663caf39e06633721c8d4cfcd7e/README.md}{GPT-2 tokenizer} revision \texttt{607a30d7} has MIT terms.

\href{https://huggingface.co/datasets/HuggingFaceFW/fineweb-edu/blob/87f09149ef4734204d70ed1d046ddc9ca3f2b8f9/README.md}{FineWeb-Edu} revision \texttt{87f09149} uses ODC-By 1.0 and Common Crawl terms. Training draws from \texttt{sample/10BT}; fresh evaluation draws from \texttt{sample/100BT}. The supervision study excludes 5,324,687 recorded document hashes. Each document receives end-of-text token 50,256. Windows contain 1,025 tokens with stride 1,024. Context may cross documents within a block but not evaluation-block boundaries. Historical development and unused test splits have 2,067/2,017 documents and 2,103,397/2,097,164 tokens; their roles remain separate.

\href{https://huggingface.co/datasets/Salesforce/wikitext/blob/b08601e04326c79dfdd32d625aee71d232d685c3/README.md}{WikiText-103 raw v1} revision \texttt{b08601e0} has 1,801,350/3,760/4,358 train/validation/test rows. The evaluated test yields 283,349 packed tokens. Its card lists CC BY-SA 3.0/GFDL in metadata and links CC BY-SA 4.0 in prose. Packed-token NLL is not word-level perplexity.

\subsection{Software and Training Details}
\label{app:software}
Task train/validation/test counts are 2,251/570/2,376 for ARC-Easy, 16,113/1,838/3,084 for PIQA, and 39,905/10,042/10,003 for HellaSwag. Pinned revisions are \texttt{210d026f}, \texttt{142f6d73}, and \texttt{218ec52e}. ARC and HellaSwag cards specify CC BY-SA 4.0 and MIT. The PIQA mirror supplies no license card; the original lists unknown terms.

The harness is 0.4.14.dev0, commit \texttt{d6de81643928}, with zero-shot official templates, no chat template, and uncached full-option likelihood. The environment uses Python 3.11.14, PyTorch 2.10.0+cu128, CUDA 12.8, Transformers 4.57.1, Tokenizers 0.22.2, Triton 3.6.0, and NumPy 1.26.4. TF32 is disabled. Muon revision \texttt{f98f1ca} is MIT-licensed; PyTorch/NumPy use BSD-style, Transformers/Tokenizers Apache 2.0, and Triton MIT terms.

Parent training has 256 warmup updates, a plateau through 6,144, and cosine decay to 10\% at 8,192, with peak rates 0.04/0.0006. C/D continue at 0.004/0.00006 to 32,768 updates, processing 1,073,741,824 targets each. GPUs use default frequency scaling and the existing 575\,W limit.

\subsection{Additional Configurations}
The label-weight and receiving-path studies retain the main architecture, batch size, sequence length, optimizer groups, and inherited rates in Table~\ref{tab:config}. They use eight RTX 5090 32\,GB GPUs, one process per GPU. The deeper configuration applies the four-block middle group three times, giving eight single-pass and sixteen looped block applications. It uses C1, C2, D1, and D2 with the original fixed loss coefficients.

Native Parcae uses a separate 32,768-token vocabulary, tied embeddings, width 768, MLP size 3,072, six attention and KV heads, and two blocks each in the prelude, recurrent core, and coda. Its core executes once in the parent and four times in the looped path. The configuration includes ReLU$^2$, QK normalization, rotary base 50,000, diagonal input injection, value embeddings, and the native random-state initialization. These changes define a separate configuration, not a tokenizer-only comparison.

Native parent seeds are 47001 and 47002; adaptation-order seeds are 47101--47106. Parents train for 32,768 updates. Both optimizer learning rates are 0.008 through update 16,384 and decay linearly to 0.0008 at update 32,768. Muon weight decay decreases from 0.2 to 0.02 over the parent run. Adaptation inherits the actual endpoint groups without warmup, including Muon momentum 0.95 with five Newton--Schulz steps and AdamW $\beta=(0.8,0.95)$, $\epsilon=10^{-10}$, and zero weight decay. Batch size is 32, microbatch size 4, accumulation 8, and sequence length 1,024, with FP32 weights and optimizer state, BF16 computation, and no activation checkpointing.

Native initialization noise is keyed by parent, order, occurrence, and window and shared across the paired paths. The frozen teacher has an independent forward computation with the same keyed noise. The main evaluation uses noise view 0; views 1 and 2 are sensitivity views, not additional training repetitions. Appendix~\ref{app:weight-study} reports the separate data and evidence scope.

\widepage{
\section{Direct Effects and Training-Order Variation}
\label{app:stats}
\begin{minipage}{\textwidth}\centering\small
\setlength{\tabcolsep}{3pt}
\resizebox{\textwidth}{!}{\input{tables/r43_effects_confirm.tex}}
\tablecaption{All twelve confirmation-text effects under both budgets. Negative favors the first method. Adjusted intervals are 98.75\% for the four output-guidance and four matched-control comparisons, and 97.5\% for the two recovery and two loop-utility comparisons. Intervals describe paired text uncertainty conditional on fitted checkpoints. Static comparisons start at the parent; staged comparisons use the respective joint suffixes.}
\label{tab:direct}
\end{minipage}\par\medskip
\begin{minipage}{\textwidth}\centering\small
\resizebox{\textwidth}{!}{\input{tables/r43_orders_proxy.tex}}
\tablecaption{Data-order variation at the forward-block cap. Each parent mean and sample SD uses three paired differences. The last column is a small-sample 95\% interval for the equal-parent mean, conditional on C/D. It does not include text uncertainty or pretraining-seed variability.}
\label{tab:order-block}
\end{minipage}\par\medskip
}

\begin{figure*}[t]\centering
\includegraphics[width=\textwidth]{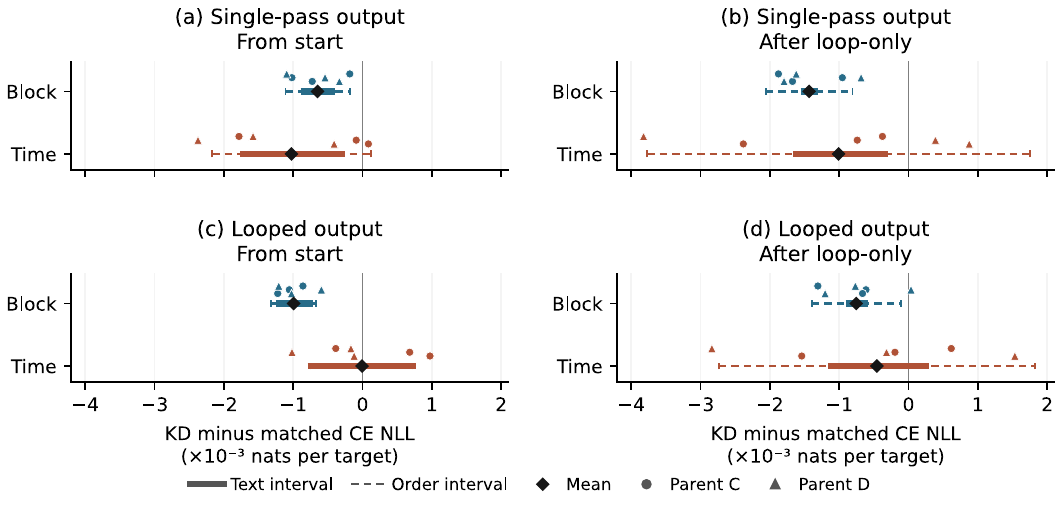}
\caption{Online KD minus coefficient-matched CE on the supervision-study confirmation text. Negative values favor KD. Columns apply the objectives from the start or after 2,048 loop-only updates; rows report the single-pass or looped output. Blue and orange denote forward-block and training-time budgets. Circles and triangles above each interval are the three paired effects within parents C and D, with orders 1--3 arranged from top to bottom within each budget. Diamonds mark the six-unit mean. Solid bars are 98.75\% adjusted text-bootstrap intervals; dashed bars are data-order 95\% intervals conditional on C/D. The intervals describe separate uncertainty sources. Values use $10^{-3}$ nats per target. Appendix~\ref{app:stats} retains all comparisons with the initial Joint CE control.}
\label{fig:staged-increment}
\end{figure*}

Text resampling uses 20,000 shared paired block draws with seed 43007. Each block contributes its summed loss and target count, and a resampled NLL divides total loss by total targets. The same block draws apply to every fitted model in a comparison, including all six continuations. The equal-parent effect averages the three paired differences within C and D, then weights the two parents equally. Averaging interval endpoints would not reproduce this interval.

The four comparison families are specified separately for each corpus and budget. They cover KD against joint CE on both outputs and schedules, joint against ordinary recovery on both outputs, the two looped improvements over ordinary continuation, and KD against matched CE on both outputs and schedules. Family sizes are 4, 2, 2, and 4. The correction applies within each family, not across every table in the paper.

Let $s_C^2$ and $s_D^2$ be sample variances of three paired data-order effects. The variance estimate for their equally weighted mean is
\begin{equation}
\widehat{\operatorname{Var}}(\bar\Delta)=\frac{s_C^2}{4R}+\frac{s_D^2}{4R},\quad R=3.
\end{equation}
The interval uses a Welch--Satterthwaite $t$ approximation with degrees of freedom derived from the two terms, each estimated with $R-1$ degrees of freedom. Three repetitions per parent give limited precision.

\widepage{
\subsection{Secondary Corpus and Time-Budget Repetitions}
\begin{minipage}{\textwidth}\centering\small
\setlength{\tabcolsep}{3pt}
\resizebox{\textwidth}{!}{\input{tables/r43_effects_wikitext.tex}}
\tablecaption{The same twelve contrasts on previously inspected WikiText. Family definitions match Table~\ref{tab:direct}, but these intervals use eighteen WikiText clusters. Results retain their corpus and budget identities.}
\label{tab:wiki-direct}
\end{minipage}\par\medskip
\begin{minipage}{\textwidth}\centering\small
\resizebox{\textwidth}{!}{\input{tables/r43_orders_time.tex}}
\tablecaption{Confirmation-text paired data-order variation at the complete-update time cap. Means and sample SDs use three orders within each fixed parent. The final interval has the same conditional interpretation as Table~\ref{tab:order-block}.}
\label{tab:order-time}
\end{minipage}\par\medskip
}

The matched-control comparison illustrates the distinction between uncertainty sources. At the forward-block cap, the staged looped mean favors KD by 0.000753. Its adjusted text interval is $[-0.000905,-0.000589]$, while its 95\% data-order interval is $[-0.001395,-0.000111]$. The D2 paired difference is still positive. A conditional mean interval does not claim that every adaptation order improves.

At the time cap, the looped KD-minus-matched-CE intervals span both signs. The static text interval bounds improvement at roughly 0.00079 and deterioration at 0.00078 nats per target; the staged interval bounds improvement at roughly 0.00116 and deterioration at 0.00030. These are readings of the reported intervals, not results of a prespecified equivalence test.

Historical guidance experiments do not contribute additional repetitions to the new means. Separate historical evaluation sets remain separate evidence.

\begin{table}[h]\centering\small
\setlength{\tabcolsep}{3pt}
\input{tables/presentation_v2/r43_endpoints_confirm_time.tex}
\caption{Absolute confirmation NLL at the complete-update time cap. Values average six paired units. Update ranges include the prefix for staged methods. Every trajectory covers the cap.}
\label{tab:time-endpoints}
\end{table}

\widepage{
\section{Selection and Corpus-Dependent Retention}
\label{app:selection}
\begin{minipage}{\textwidth}\centering\small
\resizebox{\textwidth}{!}{\input{tables/r43_wiki_successes.tex}}
\tablecaption{Every WikiText endpoint satisfying all four separate pointwise 95\% criteria. The displayed differences are point estimates; inclusion requires all four interval criteria. Diagnostic checkpoints remain infeasible on validation even when their WikiText scores satisfy the criteria. Some fixed and diagnostic rows identify the same checkpoint.}
\label{tab:wiki-success}
\end{minipage}\par\medskip
\begin{minipage}{\textwidth}\centering\small
\input{tables/presentation_v2/r43_criterion_counts.tex}
\tablecaption{Confirmation-text criteria at fixed forward-block endpoints. Each cell gives supported/violated/unresolved counts among six paired units. No method satisfies all four criteria. A failed conjunction does not imply that every criterion is violated.}
\label{tab:criteria}
\end{minipage}\par\medskip
}

\begin{table}[t]\centering\small
\input{tables/r43_selection_counts.tex}
\caption{Number satisfying all four separate pointwise 95\% criteria. Each cell evaluates 60 correlated method--unit endpoints. Diagnostic checkpoints were infeasible on validation and are not successful selections. The four intervals do not provide joint 95\% coverage.}
\label{tab:selection}
\end{table}

Selection uses the predetermined grid plus the actual time-cap state, restricted to the active budget. Other protected-tail states are excluded. The four validation point-estimate conditions are $D_{\rm start}\leq0.005$, $x\leq0.005$, $y<0$, and $d<0$. The reference is always the ordinary fixed maximum-budget endpoint in the same parent and order. Ordinary is not separately optimized by validation selection.

Among feasible states, the rule minimizes looped validation NLL, then $x$, active update seconds, and checkpoint hash. If none is feasible, the diagnostic minimizes the largest normalized violation among $(D_{\rm start}-.005)/.005$, $(x-.005)/.005$, $y/.005$, and $d/.005$, followed by the same tie-breaking rule. All 120 selections are infeasible.

The fixed and diagnostic confirmation results are each 0/60 under both budgets. Excluding ordinary gives 0/54 in each group. These groups reuse trajectories and can reuse checkpoints; they are not independent replications. The four evaluation intervals are pointwise 95\% intervals, not a simultaneous confidence region. Passing counts on WikiText are corpus-specific secondary observations and do not change the validation result.

\widepage{
\section{Task Results and Correctness Transitions}
\label{app:tasks}
\begin{minipage}{\textwidth}\centering\small
\setlength{\tabcolsep}{3pt}
\resizebox{\textwidth}{!}{\input{tables/presentation_v2/r43_tasks_primary_accuracy_both.tex}}
\tablecaption{Task accuracy (\%) in the supervision study. Each cell gives the forward-block / training-time result. Values are descriptive means of six paired continuations. ARC-Easy and PIQA use accuracy; HellaSwag uses normalized accuracy. The ordinary looped output is an unadapted diagnostic. These means do not provide pooled confidence intervals.}
\label{tab:tasks}\label{tab:task-time}
\end{minipage}\par\medskip
\begin{minipage}{\textwidth}\centering\small
\setlength{\tabcolsep}{3pt}
\resizebox{\textwidth}{!}{\input{tables/presentation_v2/r43_tasks_E_common_S0_both.tex}}
\tablecaption{Excess loss $E$ of the parent's correct examples, in percentage points. Each cell gives the forward-block / training-time result against ordinary continuation under the same budget. Negative favors the candidate. Each task and parent uses the same set $S_0$ for every method and both paths. Values are descriptive six-unit means.}
\label{tab:task-E-block}\label{tab:task-E-time}
\end{minipage}\par\medskip
}

\subsection{Evaluation and Per-Task Comparisons}
We evaluate both paths on ARC-Easy, PIQA, and HellaSwag \citep{clark2018arc,bisk2020piqa,zellers2019hellaswag}. Evaluation uses 2,376 test, 1,838 validation, and 10,042 validation items, respectively, with zero-shot official harness templates and uncached full-option likelihood. ARC-Easy and PIQA use accuracy; HellaSwag uses the harness length-normalized accuracy. Final task splits never select checkpoints or hyperparameters.

Mean accuracy can hide losses on previously correct examples. For each parent, let $S_0$ be the examples answered correctly by its original single pass. Let $F_m(S_0)$ be the fraction of $S_0$ missed by method $m$. We report the excess loss of correct predictions
\begin{equation}
E_m=F_m(S_0)-F_{\rm ordinary}(S_0).
\end{equation}
Both fractions use the same $S_0$ and the same budget's ordinary reference. Negative $E_m$ favors the candidate. Appendix~\ref{app:tasks} gives all methods, both budgets, and item-level uncertainty.

The task results are mixed (Table~\ref{tab:tasks}). Staged KD's single pass improves ARC-Easy accuracy over ordinary continuation by 1.333 percentage points (pp), with $E=-1.704$ pp. The same output changes PIQA by $-0.091$ pp and HellaSwag by $-0.037$ pp. Its looped output changes the three task accuracies by $+1.340$, $-0.299$, and $-0.053$ pp. The ARC-Easy mean improvement is therefore not a uniform task gain.

Loop-only training further separates average accuracy from retention. Its single pass changes HellaSwag accuracy by $+0.106$ pp, yet loses 1.941 pp more of the parent's correct examples than ordinary continuation. Parent HellaSwag accuracies are 26.9\% and 27.4\%, with a 25\% uniform-choice reference. The six-run means are descriptive, and the task comparisons retain their individual accuracy and retention measures.

Model updates can introduce errors despite higher mean accuracy \citep{xie2021regression}. Think-at-Hard studies correct predictions revised into errors by additional latent iterations \citep{fu2025think}. Our fixed-path comparisons instead measure adaptation-induced losses against ordinary continued training.

The common original-correct set sizes for C/D are 1,043/1,024 on ARC-Easy, 1,062/1,081 on PIQA, and 2,699/2,756 on HellaSwag. They are defined by each parent's single pass. Using a candidate-specific correct set would change the denominator and obscure retention differences.

Task uncertainty uses 20,000 paired item resamples per task with seed 43008. Item intervals condition on fitted checkpoints. Pointwise item intervals and conditional order intervals remain separate. The tables do not create a pooled interval by averaging individual bounds.

\widepage{
\section{Output and Local Update Diagnostics}
\label{app:diagnostics}
\begin{minipage}{\textwidth}\centering\small
\input{tables/presentation_v2/r43_diagnostics.tex}
\tablecaption{Diagnostic means at forward-block endpoints on eight fixed supervision-study validation blocks, comprising 262,144 targets. Fix is the fraction of tokens where looped top-1 is correct and single-pass top-1 is wrong. Break is the opposite event. These are token prediction rates, not task accuracy or sequence-level reasoning measures.}
\label{tab:diagnostics}
\end{minipage}\par\medskip
}

\subsection{What the Label Control Matches}
For a valid target $y$, temperature 1, and single-pass logits $z_s$, the single-path contribution of KD is
\begin{align}
\nabla_{z_s}\mathcal L_{\rm KD}
 &=0.1(p_s-e_y)+0.1(p_s-p_c)\notag\\
 &=0.2p_s-0.1e_y-0.1p_c.
\end{align}
Matched CE contributes $0.2p_s-0.2e_y$. Both retain $0.9\mathcal L_c$. The computation matches a nominal coefficient before clipping and optimizer transformations. Because the paths share weights, full parameter gradients also include the looped label term and contributions through the shared prefix. The algebra is not an equality of full updates.

The local diagnostic restores fourteen fixed states, comprising two original parents, six loop-only states after 2,048 updates, and six joint CE suffix states after 256 suffix updates. CE, matched CE, and KD each receive one preview update on the same fixed engineering batch, for 42 updates. The fixed batch contains 32 windows and is not a selected training or evaluation batch.

Across the fourteen states, mean clipping coefficients are 0.831672 for CE, 0.799955 for KD, and 0.741951 for matched CE. The independently reconstructed parameter-update $L_2$ norms for Muon group 0 are 0.871143, 0.871155, and 0.871159, respectively. Similar update magnitudes in one parameter group do not establish equal directions or trajectories. The diagnostic covers 126 parameter-group updates.

\subsection{Scope of the Update Diagnostics}
Muon can modify gradient buffers during the optimizer step. Norms measured afterward therefore describe the modified buffers, not the gradients immediately after clipping. The parameter-update norms above are computed from parameter differences before and after the preview update. Component gradient angles are diagnostic observations and are not used as causal evidence.

\subsection{Agreement and Prediction Quality}
Static forward KL is 0.036555 for CE, 0.026262 for KD, and 0.026378 for matched CE (Table~\ref{tab:diagnostics}). Staged forward KL is 0.045731 for CE, 0.035060 for KD, and 0.035759 for matched CE. KD fixes 1.452\% of tokens and breaks 1.266\%; matched CE fixes 1.453\% and breaks 1.270\%. The close means are descriptive and do not establish equivalence. The KD single-pass NLL is 3.636666 on this diagnostic subset, compared with 3.625906 for ordinary continuation. Improved agreement does not remove the ordinary reference gap.

The staged objectives retain a larger looped advantage within the same model than the static objectives. Their differing histories, label coefficients, and optimizer states prevent assigning the difference solely to teacher maturity. 

\widepage{
\section{Computational Resources}
\label{app:resources}
\begin{minipage}{\textwidth}\centering\small
\input{tables/presentation_v2/r43_timing.tex}
\tablecaption{Complete-update timing in the isolated RTX 5090 benchmark for the supervision study. Each method uses twenty warmup updates and five blocks of one hundred timed updates on the same device. Memory is peak allocated/reserved GiB. Measurements use default frequency scaling and the existing 575\,W limit.}
\label{tab:timing}
\end{minipage}\par\medskip
\begin{minipage}{\textwidth}\centering\small
\begin{tabular}{lr}
\toprule
Supervision-study process category & Observed GPU-process hours \\
\midrule
Training and engineering & 36.978048 \\
Validation NLL and diagnostics & 4.256758 \\
Confirmation and WikiText & 2.096609 \\
Task evaluation & 24.670267 \\
Task-interface checks & 0.004445 \\
\midrule
Total observed & 68.006127 \\
\bottomrule
\end{tabular}
\tablecaption{Approximate compute for the supervision study, including exploratory diagnostics and engineering work. Process lifetime includes overhead and is distinct from active update time. The subtotal excludes inherited parent pretraining and historical experiments.}
\label{tab:cost}
\end{minipage}\par\medskip
}

The supervision study performs 154,946 scientific updates, including 200 time-coverage updates, and 4,126 engineering updates. There are 58 gradient-only passes. Scientific input-target presentations total 5,077,270,528; engineering presentations total 135,200,768. Twenty-four staged paths reuse 49,152 logical prefix updates from the six fresh loop-only runs. The prefixes are physically trained once and are not billed again as new work. Pipeline wall time is approximately 12.44 hours.

Historical records separately report 8.896 GPU-hours of retention updates, 3.243 process hours of amplitude and earlier-continuation work, 15.235 hours of static C/D updates, 0.570 scientific and 0.192 engineering hours of ordinary recovery, and 7.512 scientific update hours for earlier guidance. Reservation bounds of 3.319 and 17.289 GPU-hours overlap other records. These records do not provide a complete pretraining and project total.

Analytic matrix-operation estimates are approximately $2.23$, $2.91$, and $4.01\times10^{13}$ FLOPs per ordinary, loop-only, and joint update. They count linear, dense attention, and Muon NS5 matrix products. They omit softmax, losses including KL, normalization, activations, recomputation, data movement, and elementwise optimizer work. Joint CE and KD therefore need measured time even when this partial FLOP estimate agrees.

Historical controls retain their original corpora and uncertainty definitions. Earlier guidance runs have one continuation per parent and condition and do not add repetitions to the present studies. Shared-prefix CE historically saves 21.2\% update time with 1.93\,GiB more allocated memory. This measurement does not include KD. Cache checks pass 48/48 FP32 and 3/48 BF16 cases under their respective tolerances; reported quality evaluations are uncached.

The target study separately records 71.956916 GPU-process hours, including training, evaluation, engineering work, and registered failures. It performs 116,706 new scientific updates and 1,792 engineering updates. Its 6,144 reused prefix updates count toward logical student exposure without a second physical charge. Frozen teachers add 52,650 full-batch forward passes. This subtotal and the supervision-study subtotal have separate coverage and exclude inherited parent pretraining.

\widepage{
\begin{minipage}{\textwidth}\centering\small
\input{tables/complete_report_20261011/q4.tex}
\tablecaption{Balanced follow-up CE timing control: four fixed GPUs, eight interleaved rounds per card. Ratios use paired card/round measurements with conditional approximate 95\% intervals. Kernel counts are not independent repetitions.}
\label{tab:completed-timing}
\end{minipage}\par\medskip
}
The label-weight and extension study records 110.113069 training-process and 29.898895 evaluation-process GPU-hours, totaling 140.011964. Including engineering work, failures, and conservatively estimated overhead gives 146.056127 GPU-process hours. Teacher computation and native parent training are included; reused evaluations are counted once and idle reservations are excluded. These are process-time estimates rather than exact device-utilization measurements. Complete CPU usage and project-wide historical GPU usage were not measured.

The common weighted-CE entry differs only in the loss coefficients and shows almost equal update times, whereas the historical separate entries reproduce an approximately 5.2\% gap (Table~\ref{tab:completed-timing}). Traces show observable implementation differences, including synchronization, but do not assign the entire gap to one operation. These measurements neither replace historical update rates nor change saved time-budget endpoints or their quality results.

\widepage{
\section{Target Content, Teacher Source, and Guidance Timing}
\label{app:targets}
\begin{minipage}{\textwidth}\centering\small
\setlength{\tabcolsep}{2pt}
\resizebox{\textwidth}{!}{\input{tables/r43v2_effects_exposure.tex}}
\tablecaption{All predefined target-study comparisons at 2,925 student updates. Values are in $10^{-3}$ nats per target. Negative favors the first method. Text intervals retain the original content/source/timing family corrections of 99.1667\%/99.375\%/99.5\%; the re-evaluated KD-minus-matched-CE row retains 98.75\%. Order intervals are conditional Welch 95\% intervals for the two fixed parents. KD denotes the current online looped target; frozen methods use the corresponding initial predictors. Horizontal rules separate the three intervention families and the re-evaluated control.}
\label{tab:target-exposure}
\end{minipage}\par\medskip
}

\subsection{Targets and Shared Training Conditions}
The initial target study uses Eq.~\eqref{eq:target} with $\alpha=0.1$, temperature 1, and full-vocabulary KL averaged over valid targets. The seven additions are frozen looped, frozen single-pass, online flat, online permuted, frozen looped permuted, early-only, and delayed guidance. Frozen teachers compute their own activations from input tokens with fixed initial weights. Online targets follow each branch's current parameters.

For vocabulary size $V=50{,}257$ and label $y$, the flat target retains $q_y$ and assigns $(1-q_y)/(V-1)$ elsewhere. Permutation preserves $q_y$, entropy, and the probability multiset. PCG64 seeds 143001/143002 produce one fixed permutation per parent, shared across its three orders and online/frozen controls. A label-preserving adjustment fixes $y$ at each position. Labels construct training targets and do not enter inference.

Early-only guidance uses KD for 512 updates, then matched CE for 2,413; delayed guidance reverses the objectives. Both restore the corresponding full 512-update state, including optimizer, random state, and data cursor. The interaction is continuous minus early-only minus delayed plus matched CE.

\subsection{Data, Repetitions, and Intervals}
Fresh FineWeb-Edu validation has 484 documents, sixteen blocks, and 524,288 targets. Confirmation has 2,057 documents, thirty-two blocks, and 2,097,152 targets. Source, packing, and historical hash exclusion follow Appendix~\ref{app:assets}. Document sets differ from the supervision study. Parents, data orders, and inherited training windows are shared. Re-evaluated controls add no training repetitions.

Text bootstrap uses 20,000 shared draws with seed 143007. Content/source/timing families contain 6/8/10 effects across both outputs. Tables retain their original corrections. Data-order intervals use the fixed-parent Welch construction in Appendix~\ref{app:stats}. Practical equivalence requires the relevant interval wholly inside $\pm0.001$ on each output. At equal updates, early-only guidance versus matched CE and continuous versus delayed guidance have both text-family and order intervals inside $\pm0.001$ on each output. Omitting the first 512 guidance updates therefore preserves the final result at this scale. Time-budget intervals do not support the same conclusion. Appendix~\ref{app:targets} retains all target contrasts and teacher diagnostics.

\widepage{
\subsection{Resource Budgets and Complete Contrasts}
\begin{minipage}{\textwidth}\centering\small
\setlength{\tabcolsep}{2pt}
\resizebox{\textwidth}{!}{\input{tables/r43v2_effects_proxy.tex}}
\tablecaption{Target-study comparisons at the 40,960-block cap including teacher computation. Units, contrasts, and intervals match Table~\ref{tab:target-exposure}. Frozen looped and single-pass endpoints have 1,575 and 1,861 student updates; online joint endpoints have 2,925. Fixed-update and fixed-resource comparisons retain separate identities.}
\label{tab:target-block}
\end{minipage}\par\medskip
\begin{minipage}{\textwidth}\centering\small
\setlength{\tabcolsep}{2pt}
\resizebox{\textwidth}{!}{\input{tables/r43v2_effects_time.tex}}
\tablecaption{Target-study comparisons at 2,227.587 seconds of complete model updates, including teacher forward passes, target construction, and reused prefixes. Units and intervals match Table~\ref{tab:target-exposure}. The time endpoints cover the cap, with the last saved state within it used for evaluation.}
\label{tab:target-time}
\end{minipage}\par\medskip
}

Joint student computation uses fourteen block applications. Frozen looped and single-pass teachers add twelve and eight, respectively. Equal-update comparisons fix student exposure, whereas Tables~\ref{tab:target-block} and \ref{tab:target-time} include the extra teacher cost. The original measured update times for CE, matched CE, and KD in Table~\ref{tab:timing} remain separate observations. The timing comparison uses those measurements without attributing their differences to a single implementation component.

\subsection{Teacher Quality and Retention References}
Teacher diagnostics use eight fixed validation blocks with 262,144 targets. Teacher-minus-student NLL is positive for the frozen initial looped teacher at every scheduled state in all six branches, with pointwise 95\% intervals supporting that direction. The online looped teacher is above its single-pass student at initialization and below it at measured updates 128, 256, and 512. These are discrete, pointwise comparisons. Some matched CE and ordinary reference states are missing or outside the legal budget.

Prediction-permutation diagnostics have an established precedent \citep{furlanello2018born}. Permuting a frozen looped target's non-label probabilities preserves its gold-token probability, entropy, and probability multiset. At equal updates, the unpermuted target lowers single-pass NLL by 0.463495 relative to the permuted target, yet has 0.028096 higher NLL than matched CE. The intervention shows sensitivity to probability-to-token correspondence. It does not identify the source of the much smaller online-distillation residual.

The target study selects from 1,790 legal validation entries across seventeen methods, six units, and two budgets. None of the 204 selection problems has a feasible state under the four criteria in \S\ref{sec:setup}. Each retains the predefined minimum-violation diagnostic state. A separate strong ordinary reference minimizes ordinary validation NLL within the budget, breaking ties by cost, update count, and checkpoint hash. Fixed endpoints compared with fixed and validation-selected ordinary references give 408 correlated confirmation comparisons; none meets all four separate pointwise 95\% criteria. These counts are separate from the supervision study's 120 selection problems.

\begin{table}[t]\centering\small
\setlength{\tabcolsep}{3pt}
\resizebox{\columnwidth}{!}{\input{tables/r43v2_retention_time.tex}}
\caption{Target-study mean NLL differences at the time cap against validation-selected ordinary continuation. Definitions follow \S\ref{sec:setup}. Each entry averages six paired units; the means are not new significance tests.}
\label{tab:target-retention}
\end{table}

Under the strong ordinary reference, the frozen single-pass teacher's mean single-pass gap is 0.012772 at the time cap (Table~\ref{tab:target-retention}). Its comparison with matched CE therefore answers a different question from retention of ordinary continued-learning gains. Shared-weight effects can improve the looped output while leaving the single-pass reference gap.

\subsection{Task Comparisons}
Task evaluation retains the same splits, templates, and common original-correct sets as Appendix~\ref{app:tasks}. Of 303 evaluated state groups, 160 are new evaluations and 143 reuse verified original-study option scores. At the time cap against validation-selected ordinary continuation, online KD has ARC-Easy single-pass accuracy difference $+0.877$ pp, with data-order 95\% interval $[-0.892,+2.646]$, and excess loss $E=-2.477$ pp with interval $[-4.472,-0.482]$. The corresponding PIQA and HellaSwag accuracy differences are $-0.462$ and $-0.003$ pp, with intervals $[-1.479,+0.554]$ and $[-0.432,+0.425]$. Their $E$ intervals are $[-0.352,+0.887]$ and $[-0.799,+0.771]$ pp. Task and retention measures thus retain separate interpretations.

\widepage{
\subsection{Supplementary Threshold Sensitivity}
\label{app:sensitivity}
\begin{minipage}{\textwidth}\centering\small
\input{tables/r46_ppl_scales.tex}
\tablecaption{Algebraic conversion of NLL increases to relative perplexity increases, $100(\exp(\Delta\mathrm{NLL})-1)$. The conversion uses the same tokenization and evaluation protocol. It is neither a task-accuracy change nor a comparison of word-level perplexity across tokenizers.}
\label{tab:ppl-scales}
\end{minipage}\par\medskip
\begin{minipage}{\textwidth}\centering\small
\setlength{\tabcolsep}{4pt}
\input{tables/r46_retention_scales.tex}
\tablecaption{Supplementary retention sensitivity on each study's primary confirmation text. Cells count supported/violated/unresolved conjunctions among fixed endpoints at the column's tolerance $\tau$. The requirements are $D_{\rm start}\leq\tau$, $x\leq\tau$, $y<0$, and $d<0$. Supervision rows contain 60 method--unit endpoints; target rows contain 102. Fixed and selected denote the ordinary reference, not candidate selection. Selected ordinary states minimize validation NLL within the budget. Each conjunction uses the four separate pointwise 95\% intervals for the same checkpoint and budget. The counts are correlated and do not provide simultaneous coverage.}
\label{tab:retention-scales}
\end{minipage}\par\medskip
\begin{minipage}{\textwidth}\centering\small
\input{tables/r46_equivalence_scales.tex}
\tablecaption{Single-pass equivalence sensitivity for guidance from the frozen original single pass minus matched CE. Columns give the half-width of the equivalence range in nats per target. S means the entire interval is inside the range, V means it is entirely outside, and U means it overlaps a boundary. Looped intervals lie below every tested equivalence range, reflecting improvements beyond these scales. Text-family and data-order intervals remain separate.}
\label{tab:equivalence-scales}
\end{minipage}\par\medskip
}

This supplementary analysis reclassifies archived intervals after the experiments. It does not change the original 0.005 retention tolerance, the target study's 0.001 equivalence scale, or the selection rules. No models are retrained, weights retuned, checkpoints reselected, or intervals refitted. The margins describe NLL scales rather than deployment requirements or established boundaries for downstream capability.

For retention, an interval supports a criterion when its upper bound meets the threshold and violates it when its lower bound exceeds the threshold; the zero-gain conditions use strict inequalities. A conjunction is supported when all four criteria are supported, violated when at least one is violated, and unresolved otherwise. All four effects belong to the same candidate and budget. Varying the tolerance changes the accepted single-pass cost. Both looped-gain conditions remain fixed.

For equivalence, the entire interval must lie within $[-\epsilon,\epsilon]$. An interval wholly outside the range violates equivalence, and any remaining case is unresolved. The frozen original teacher's time-matched single-pass intervals are $[-0.000634,+0.001595]$ for adjusted text uncertainty and $[-0.000704,+0.001778]$ for data-order uncertainty. Both are contained in $\pm0.002$ but not $\pm0.001$. The comparison is with matched CE and leaves the 0.012772 mean gap to validation-selected ordinary continuation unchanged.

\clearpage
\begin{figure*}[t]\centering
\includegraphics[width=\textwidth]{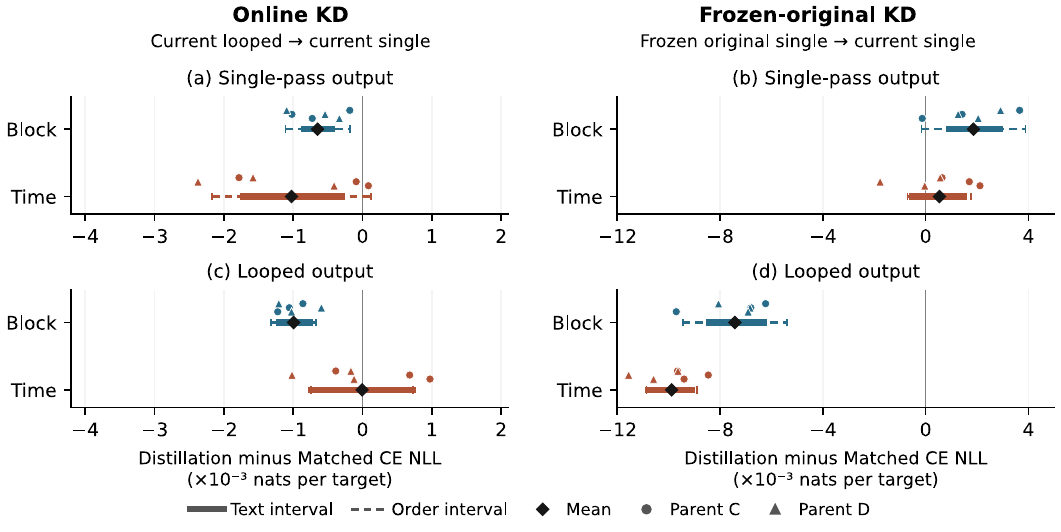}
\caption{Distillation effects beyond stronger single-pass labels. Online KD leaves a small residual, whereas Frozen-original KD primarily improves the looped output. Both methods start at the original checkpoint. All KL terms guide the current single pass; both label losses remain active. Each column compares with Matched CE on its own study's confirmation text, so the columns do not rank methods across datasets. Columns use different horizontal ranges. Negative values favor distillation. Blue Block and orange Time rows use forward-block and complete-update time budgets, including teacher computation. Circles and triangles show three paired effects within parents C and D, with orders 1--3 from top to bottom. Diamonds mark the six-unit mean. Solid bars are adjusted text-bootstrap intervals, 98.75\% for Online KD and 99.375\% for Frozen-original KD. Dashed bars are data-order 95\% intervals conditional on C/D. The two interval types describe separate uncertainty sources. Values use $10^{-3}$ nats per target. Appendix Figure~\ref{fig:staged-increment} retains both online-KD schedules.}
\label{fig:paired-teachers-full}
\end{figure*}

\clearpage
\begin{figure*}[t]\centering
\includegraphics[width=\textwidth]{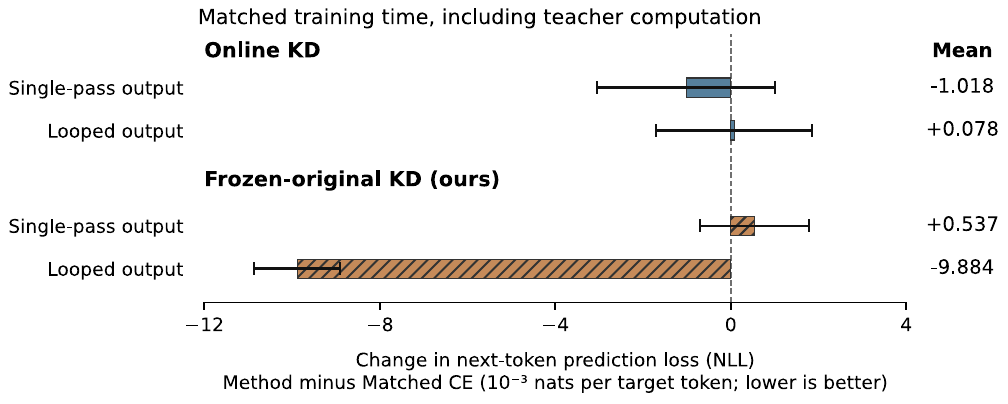}
\caption{Initial fixed-coefficient target study. \textbf{Guidance to the single pass mainly improves the looped output with Frozen-original KD (ours).} Bars show NLL differences from Matched CE at matched training time, including teacher computation, on the same target-study confirmation text. Lower is better. Both methods guide only the current single pass and retain label losses on both outputs. Error bars are 95\% data-order intervals for six paired continuations, conditional on two fixed pretrained models. Adjusted text intervals, block-budget results, and individual effects are reported in Appendix~\ref{app:targets}.}
\label{fig:fixed-teacher-time}
\end{figure*}

\begin{table*}[t]\centering\small
\setlength{\tabcolsep}{4pt}
\input{tables/presentation_v2/minimal_teacher_time.tex}
\caption{Initial fixed-coefficient target study. Teacher-source effects at the complete-update time cap, in $10^{-3}$ nats per target. Every distillation objective guides the current single pass while retaining label losses on both current outputs. Differences are candidate minus coefficient-matched CE on the target-study confirmation text; negative favors the candidate. Teacher computation is included. Means summarize six paired continuations, and 95\% intervals describe data-order variation conditional on the two pretrained parents. The online row re-evaluates existing trajectories. Equal-update and text-bootstrap results appear in Appendix~\ref{app:targets}.}
\label{tab:fixed-teacher-time}
\end{table*}

\widepage{
\section{Label-Weight Selection and Additional Experiments}
\label{app:weight-study}
\subsection{Selection Rule and Complete Finite Grid}
\begin{minipage}{\textwidth}\centering\small
\input{tables/new_results_20261011/selection.tex}
\tablecaption{Weights selected from validation data on C1 and D1. The primary tolerance is $\tau=0.005$; other rows are the prespecified development-set sensitivity analysis. None means no admissible weight, not a missing evaluation or a substituted recipe. The frozen-teacher family selects $m=0.4$ throughout.}
\label{tab:new-selection}
\end{minipage}\par\medskip
\begin{minipage}{\textwidth}\centering\small
\setlength{\tabcolsep}{4pt}
\input{tables/new_results_20261011/full_grid.tex}
\tablecaption{All twelve measured weight-grid points at the complete-update time cap on the new confirmation text. Values are descriptive six-order means in nats per target and include the development orders C1 and D1. They are separate from the four-order selected-recipe confirmation in Table~\ref{tab:target-primary}. All four counts orders meeting the four point-estimate criteria with the primary tolerance. No untrained intermediate weights or continuous frontier are inferred.}
\label{tab:new-grid}
\end{minipage}\par\medskip
}

The three method families use the same finite grid in Eq.~\ref{eq:grid}. For each family, a weight is admissible when both C1 and D1 satisfy $D_{\rm start}\leq0.005$ and $x\leq\tau$ on validation at the fixed-time endpoint. Among admissible weights, selection minimizes their mean looped validation NLL. Exact ties prefer the lower $m$, then checkpoint hash. An infeasible family receives no selected weight. The looped conditions $y<0$ and $d<0$ are evaluated after selection; they do not enter eligibility. Confirmation text is opened after the global weight choices are frozen.

At the primary tolerance, label-only CE uses $0.8\mathcal L_s+0.9\mathcal L_c$. Online KD uses $0.7\mathcal L_s+0.9\mathcal L_c+0.1\operatorname{KL}(\operatorname{sg}[p_c]\|p_s)$. Frozen-original KD uses the same KL coefficient with $q=p_s(\theta_0)$ and single-pass label coefficient 0.3. All coefficients are unnormalized. The selected CE recipe is different from the initial coefficient-matched CE control with weight 0.2.

Confirmation means use C2, C3, D2, and D3, weighting the two orders within each parent equally and then the parents equally. These orders are held out from the present weight-selection procedure. C/D and historical low-weight trajectories had prior research use; the procedure does not make them previously unseen pretrained models or training orders.

The full grid illustrates a trade-off. The low frozen-teacher weight $m=0.2$ retains a looped advantage over ordinary continuation but has a larger single-pass gap. At $m=0.4$, all six descriptive endpoints meet the criteria. At $m=0.8$, the looped output no longer improves on the same model's single pass on average; at $m=1.6$, its mean NLL exceeds ordinary continuation. These are measured-point observations, not a global optimality or dominance test.

\widepage{
\begin{minipage}{\textwidth}\centering\small
\input{tables/complete_report_20261011/grid_comparisons.tex}
\tablecaption{Supplementary paired comparisons of frozen $m=0.4$ with every tested CE weight, at the time cap on confirmation text. Units are $10^{-3}$ nats per target; negative favors the frozen teacher. These six-order means include the development orders and are distinct from the four-order primary confirmation. Text intervals retain the 132-effect correction across all measured-point pairs and both coordinates in this budget/text/summary view.}
\label{tab:completed-grid}
\end{minipage}\par\medskip
}
This supplementary analysis was specified after the preliminary outcomes had been examined. It compares only twelve trained points, without interpolation or additional selection. Of the 66 pairs at the primary time cap, 17 have a supported strict two-coordinate relationship under the adjusted text intervals; these are not 17 independent experiments. Table~\ref{tab:completed-grid} verifies the favorable direction for frozen $m=0.4$ against each of the four CE weights. The 20,000 paired bootstrap draws leave only about 3.79 draws in each extreme tail of the 132-effect family, limiting Monte Carlo precision. The correction is within each budget/text/summary view, not across all views or both text and order uncertainty. A relationship not established is not equivalence. WikiText has one scoring block and remains descriptive.

\widepage{
\subsection{Uncertainty for the Selected Recipes}
\begin{minipage}{\textwidth}\centering\small
\setlength{\tabcolsep}{3pt}
\input{tables/new_results_20261011/selected_intervals.tex}
\tablecaption{Selected-recipe criteria on the new confirmation text, in nats per target. Each estimate uses the same four held-out orders. Text intervals use the prespecified twelve-effect family (three recipes times four criteria), giving 99.5833\% marginal coverage. Order intervals are separate conditional Welch 95\% intervals for the two fixed parents, each with two orders. The thresholds are 0.005 for $D_{\rm start}$ and $x$, and strictly zero for $y$ and $d$. All four adjusted text upper bounds for Frozen-original KD meet their thresholds. The interval for $x$ crosses zero, so the conclusion is retention within the margin.}
\label{tab:new-intervals}
\end{minipage}\par\medskip
\begin{minipage}{\textwidth}\centering\small
\setlength{\tabcolsep}{3pt}
\input{tables/new_results_20261011/selected_effects.tex}
\tablecaption{Direct effects of selected KD recipes relative to selected CE, using the same four orders and fixed-time endpoints. Negative favors the KD recipe. The four comparisons form a separate family with adjusted 98.75\% text intervals. Fixed-parent order intervals remain pointwise 95\%. No interval combines text variation with variation across continuation orders or pretrained initializations.}
\label{tab:new-effects}
\end{minipage}\par\medskip
}

Text uncertainty uses 20,000 shared paired resamples with seed 47007. Resampling sums block losses and divides by the corresponding total target count before averaging paired order effects. Intervals use full-precision paired differences rather than rounded means or averaged individual bounds. Text intervals condition on the fitted checkpoints. The two orders per parent give conditional order intervals using the Welch--Satterthwaite construction in Appendix~\ref{app:stats}, with $R=2$.

\paragraph{Statistical checks.} The text and task statistics were reconstructed from saved model outputs. The reconstruction shares the checked numerical kernel with the original analysis and is not an independent implementation of the bootstrap.

The four Frozen-original KD point estimates meet the criteria in every held-out order (Table~\ref{tab:new-orders}). This count is not a simultaneous per-order confidence guarantee. The corrected intervals in Table~\ref{tab:new-intervals} instead support the four-order mean under the prespecified text family. The separate order intervals meet the thresholds but are not jointly adjusted by the text-family procedure.

Frozen-original KD improves looped NLL over selected CE in each of the four orders. Its single-pass difference is positive in D3 and negative in the other three orders. Online KD has positive mean differences on both outputs relative to selected CE, but its conditional order intervals span zero. These observations do not imply the same ordering for every future continuation.

\widepage{
\subsection{Individual Orders, Data, and Budget Identity}
\begin{minipage}{\textwidth}\centering\small
\setlength{\tabcolsep}{3pt}
\input{tables/new_results_20261011/selected_orders.tex}
\tablecaption{All individual selected-recipe effects on the new primary confirmation text. Values are in nats per target. C2/C3 and D2/D3 identify the two retained orders for each fixed parent. Each row uses one candidate checkpoint for all four criteria and its paired fixed-time ordinary reference. Yes denotes joint satisfaction of the point-estimate inequalities, without an interval-based per-run claim.}
\label{tab:new-orders}
\end{minipage}\par\medskip
\begin{minipage}{\textwidth}\centering\small
\begin{tabular}{llrrr}
\toprule
Configuration & Split & Documents & Targets & Scoring blocks \\
\midrule
Main / deeper, GPT-2 tokenizer & Validation & 524 & 524,288 & 16 \\
Main / deeper, GPT-2 tokenizer & Confirmation & 2,175 & 2,097,152 & 32 \\
Main / deeper, GPT-2 tokenizer & WikiText & -- & 282,624 & 1 \\
Native, Parcae tokenizer & Validation & 524 & 524,288 & 16 \\
Native, Parcae tokenizer & Confirmation & 2,175 & 2,125,824 & 32 \\
\bottomrule
\end{tabular}
\tablecaption{New-study text scoring views. Native evaluation retokenizes the same document blocks, producing different token counts. Absolute NLL is not compared across tokenizers. WikiText is a previously inspected secondary corpus with one scoring block in this study; the earlier eighteen-cluster view and its intervals are not reused.}
\label{tab:new-data}
\end{minipage}\par\medskip
}

New validation and confirmation text exclude recorded training and earlier evaluation documents by exact document hash. Validation selects weights; confirmation text does not. The primary reference is ordinary single-pass continuation from the same parent and order at its fixed complete-update time endpoint.

The cap is $T\leq2,227.587$ seconds. It includes teacher forward computation, all losses, backpropagation, clipping, optimizer updates, and synchronization. Each endpoint is the last actual saved state within the cap, without interpolation or a claim that every method exhausts exactly the same number of seconds. Loading, evaluation, and checkpoint writing enter the separate process-lifetime ledger. Protected-tail updates establish budget coverage; states beyond the cap are not substituted into the primary comparison. Other equal-update, forward-block, and within-budget checkpoint-selection views remain separate.

The main selected-recipe table uses fixed-time ordinary continuation. The older 0.012772 single-pass gap in Table~\ref{tab:target-retention} uses a different recipe, confirmation text, six-order average, and validation-selected ordinary reference. It remains an observation from the initial target study and cannot be transferred to the new recipe.

A separate within-budget analysis selects checkpoints as follows: both candidate and ordinary minimize validation single-pass NLL over predefined saved states with step no greater than the primary exposure endpoint and active time no greater than $T$, including initialization. Ties prefer fewer updates, then checkpoint hash. The ordinary pool ends at 2,925 updates, excluding later time-coverage states; the selected ordinary steps for C2/C3/D2/D3 are 2,925/2,925/2,560/2,925. These are different checkpoints from the primary fixed-time references. Existing aggregate intervals use all six orders and also select the candidate checkpoint. They therefore do not establish the four-held-out-order comparison of the fixed final recipe against a stronger validation-selected ordinary reference. No such interval is inferred from the old study or from differently scoped means; this limitation remains.

\widepage{
\subsection{Receiving Path and Configuration Extensions}
\begin{minipage}{\textwidth}\centering\small
\setlength{\tabcolsep}{4pt}
\input{tables/new_results_20261011/receivers.tex}
\tablecaption{Frozen-teacher guidance minus its respective coefficient-matched label control, in nats per target. Values are descriptive six-order means at matched training time on the new confirmation text. Every condition uses the frozen original single-pass teacher and total KL weight 0.1. Guidance to both outputs splits the weight into 0.05 per output. The controls differ across rows, so the improvement magnitudes do not rank receiving paths.}
\label{tab:new-receivers}
\end{minipage}\par\medskip
\begin{minipage}{\textwidth}\centering\small
\setlength{\tabcolsep}{3pt}
\input{tables/new_results_20261011/extensions.tex}
\tablecaption{Frozen-original KD in separate configurations at the time cap. $x$ and $y$ are differences from each configuration's own ordinary single-pass reference; the final numerical column compares looped outputs with the stated CE recipe. Values are descriptive means in nats per target; $n$ counts parent--order units. The WikiText row does not reconstruct a paired CE effect from differently scoped averages. The deeper and native rows use the original fixed coefficients, not the selected main recipe. These within-configuration differences do not compare absolute NLL across vocabularies.}
\label{tab:new-extensions}
\end{minipage}\par\medskip
}

The receiving-path study uses the original label weights $(0.1,0.9)$ and total KL weight 0.1. Its label controls match the nominal coefficient for each receiving output, giving $(0.2,0.9)$ for single-pass guidance, $(0.1,1.0)$ for looped guidance, and $(0.15,0.95)$ for guidance to both. Direct looped guidance improves mean single-pass NLL over its own label control. Guidance need not target the single pass to benefit that output. The completed paired comparisons in Table~\ref{tab:completed-receivers} separate direct recipe differences from interactions with their respective label controls. Loop-only guidance has lower looped NLL than guidance to both outputs under the adjusted text comparison, but the conditional order interval spans zero. These original-coefficient interventions do not identify the best receiving path for the validation-selected recipe.

The loop-guided and both-guided recipes have mean single-pass gaps $x=0.009620$ and $0.010181$, and looped gaps $y=-0.013773$ and $-0.011857$. Each has zero of six endpoints meeting all four point-estimate criteria at the original coefficients. These comparisons do not replace the main label-weight selection.

The deeper configuration uses four orders C1, C2, D1, and D2. Frozen-original KD has single-pass and looped differences of $+0.003189$ and $-0.010057$ relative to fixed CE. Native Parcae uses two new R1 parents and three adaptation orders per parent. Its corresponding differences are $-0.002250$ and $-0.002413$. Neither frozen-teacher extension satisfies all four point-estimate criteria at matched training time. The studies evaluate transfer of the original fixed recipe, without selecting $m=0.4$ for these configurations.

For the selected main recipe on WikiText, $d=-0.008728$, so looping improves the same adapted model's NLL. Both $x$ and $y$ remain positive relative to ordinary continuation. Selected CE and Online KD have looped gaps $y=0.028427$ and $0.027315$. The one-block scoring view cannot provide document-level bootstrap uncertainty. The main primary-text result therefore retains its corpus scope.

\widepage{
\begin{minipage}{\textwidth}\centering\small
\setlength{\tabcolsep}{3pt}
\input{tables/complete_report_20261011/receiver_interactions.tex}
\tablecaption{Completed receiving-path comparisons, in $10^{-3}$ nats per target, for six orders at the time cap on confirmation text. Direct is KD$_A-$KD$_B$; interaction is (KD$_A-$CE$_A$)$-$(KD$_B-$CE$_B$). CE controls are receiver-specific. All twelve comparisons retain their 99.5833\% text-family intervals; conditional order 95\% intervals are separate. The original label coefficients remain fixed.}
\label{tab:completed-receivers}
\end{minipage}\par\medskip
\begin{minipage}{\textwidth}\centering\small
\setlength{\tabcolsep}{3pt}
\input{tables/complete_report_20261011/structure_intervals.tex}
\tablecaption{Fixed-recipe extension intervals on each configuration's confirmation text, in $10^{-3}$ nats per target. HF uses four orders; native uses six, with primary evaluation noise view 0. Text intervals retain each configuration's six-effect family. These are not tests of the selected main coefficient $\alpha=0.3$, nor comparisons of absolute NLL across tokenizers.}
\label{tab:completed-structure}
\end{minipage}\par\medskip
}
The deeper configuration has an adverse single-pass text interval despite its looped gain; its single-pass order interval spans zero. Native frozen guidance improves over its matched CE control but still has higher NLL than ordinary continuation on both paths. The architecture, tokenizer, and training recipe change together in the native extension. No second-setting joint success of the selected main recipe is established.

\begin{table}[h]\centering\small
\input{tables/complete_report_20261011/native_noise.tex}
\caption{Native fixed-recipe sensitivity to evaluation noise, in $10^{-3}$ nats per target. Each row averages the same six fitted models at the time cap; views 1 and 2 are not new training repetitions. All displayed $x$ and $y$ means are positive relative to native ordinary continuation.}
\label{tab:completed-noise}
\end{table}

\widepage{
\subsection{Task Accuracy and Retention}
\begin{minipage}{\textwidth}\centering\small
\setlength{\tabcolsep}{4pt}
\input{tables/new_results_20261011/tasks.tex}
\tablecaption{Task accuracy (\%) for the selected recipes and paired ordinary continuation. Every row averages the same four held-out orders at the time cap. ARC-Easy and PIQA use accuracy; HellaSwag uses normalized accuracy. The ordinary row is recomputed from the four matching orders, not the six-order reference in the initial studies. These means are descriptive and have no pooled task interval.}
\label{tab:new-tasks}
\end{minipage}\par\medskip
\begin{minipage}{\textwidth}\centering\small
\setlength{\tabcolsep}{3pt}
\input{tables/complete_report_20261011/task_retention.tex}
\tablecaption{Original-correct-set retention for all three selected recipes and both outputs on all tasks. For each parent and task, $S_0$ contains examples its original single pass answers correctly. $F$ is the error proportion within $S_0$; $E=F_{\rm candidate}-F_{\rm ordinary}$, so negative values mean less loss than ordinary, not an accuracy gain or zero forgetting. Each row averages C2/C3/D2/D3 at the fixed-time endpoints. Item intervals are paired, pointwise 95\%, conditional on fitted models; order intervals are separate fixed-parent Welch 95\% intervals. Neither gives a cross-task simultaneous guarantee.}
\label{tab:completed-tasks}
\end{minipage}\par\medskip
}

Task evaluation uses ARC-Easy test (2,376 items), PIQA validation (1,838), and HellaSwag validation (10,042), with the zero-shot templates and metrics described in Appendix~\ref{app:software}. Tasks do not select weights or checkpoints. Frozen-original KD has higher mean ARC-Easy accuracy than ordinary continuation, while its HellaSwag means are lower. The three selected recipes do not have a uniform ranking across tasks. The observed text-NLL result is not a claim of broad task improvement.

Table~\ref{tab:completed-tasks} reports the completed original-correct-set analysis. For frozen $m=0.4$, ARC-Easy excess loss is $-3.5337$ pp for the single pass (item 95\% interval $[-4.8440,-2.2635]$) and $-2.8279$ pp for the looped output ($[-4.1896,-1.5158]$). Both conditional order intervals span zero. PIQA and HellaSwag show why the favorable ARC-Easy direction is not a general task-retention result. The conditional denominator $|S_0|$ differs from negative-flip rates defined over all examples \citep{xie2021regression}, and $E$ additionally subtracts ordinary continuation's loss.

Task statistics use 20,000 paired item resamples with seed 47008. Additional task analyses consider individual orders, correct-to-incorrect and repair transitions, option margins, exact McNemar comparisons, and predefined 1/2 pp tolerance views. Endpoint intervals in Table~\ref{tab:completed-tasks} are pointwise. Separate direct-comparison corrections of 4/6/12 effects apply within each task and metric only. No task result changes a weight or checkpoint choice; item and order uncertainty do not form a joint guarantee.

\end{document}

%% file: tables/presentation_v2/r43_endpoints_confirm_proxy.tex
\begin{tabular}{lrrr}
\toprule
Method & Updates & Single & Looped \\
\midrule
Ordinary & 5120 & 3.638339 & -- \\
Loop-only & 3413 & 3.809579 & 3.629291 \\
Joint CE 5\% & 2925 & 3.674221 & 3.632287 \\
Joint CE & 2925 & 3.661175 & 3.633369 \\
Online KD & 2925 & 3.650679 & 3.634773 \\
Matched CE & 2925 & 3.651328 & 3.635767 \\
Loop$\to$CE & 3218 & 3.667589 & 3.630223 \\
Loop$\to$KD & 3218 & 3.654835 & 3.630768 \\
Loop$\to$matched CE & 3218 & 3.656269 & 3.631521 \\
Loop$\to$ordinary & 4096 & 3.639498 & 3.843826 \\
\bottomrule
\end{tabular}

%% file: tables/complete_report_20261011/selected.tex
\begin{tabular}{lrrrrc}
\toprule
Recipe & $D_{\rm start}$ & $x$ & $y$ & $d$ & All four \\
\midrule
Label-only CE, $m=0.8$ & $-6.291$ & $+1.929$ & $+3.571$ & $+1.642$ & 0/4 \\
Online KD, $m=0.8$ & $-5.657$ & $+2.563$ & $+4.641$ & $+2.078$ & 0/4 \\
Frozen-original KD, $m=0.4$ & $-8.953$ & $-0.733$ & $-8.327$ & $-7.595$ & 4/4 \\
\bottomrule
\end{tabular}

%% file: tables/complete_report_20261011/key_intervals.tex
\begin{tabular}{lrrr}
\toprule
Key contrast, Frozen-original KD & Mean & Adjusted text interval & Order 95\% \\
\midrule
Single $-$ ordinary & $-0.733$ & $[-1.940, +0.446]$ & $[-4.150, +2.684]$ \\
Looped $-$ ordinary & $-8.327$ & $[-9.513, -7.225]$ & $[-13.522, -3.133]$ \\
Looped $-$ selected CE & $-11.899$ & $[-13.084, -10.875]$ & $[-20.206, -3.592]$ \\
\bottomrule
\end{tabular}

%% file: tables/r43_effects_confirm.tex
\begin{tabular}{lrrrr}
\toprule
Comparison & Block mean & Text interval & Time mean & Text interval \\
\midrule
Static KD $-$ CE, single & $-0.010496$ & $[-0.010766,\,-0.010229]$ & $-0.010704$ & $[-0.011431,\,-0.009958]$ \\
Static KD $-$ CE, looped & $+0.001404$ & $[+0.001166,\,+0.001648]$ & $+0.002166$ & $[+0.001409,\,+0.002942]$ \\
Staged KD $-$ CE, single & $-0.012754$ & $[-0.012976,\,-0.012528]$ & $-0.011724$ & $[-0.012606,\,-0.010809]$ \\
Staged KD $-$ CE, looped & $+0.000545$ & $[+0.000406,\,+0.000686]$ & $+0.001023$ & $[+0.000082,\,+0.001982]$ \\
CE $-$ ordinary suffix, single & $+0.028091$ & $[+0.027243,\,+0.028919]$ & $+0.028047$ & $[+0.027165,\,+0.028953]$ \\
CE $-$ ordinary suffix, looped & $-0.213603$ & $[-0.215512,\,-0.211710]$ & $-0.210361$ & $[-0.211949,\,-0.208645]$ \\
Joint CE 5\% $-$ Ordinary, loop/single & $-0.006052$ & $[-0.006997,\,-0.005100]$ & $-0.005907$ & $[-0.006685,\,-0.005149]$ \\
Loop-only $-$ Ordinary, loop/single & $-0.009048$ & $[-0.010027,\,-0.008021]$ & $-0.010880$ & $[-0.011658,\,-0.010081]$ \\
Static KD $-$ matched CE, single & $-0.000649$ & $[-0.000892,\,-0.000397]$ & $-0.001025$ & $[-0.001770,\,-0.000257]$ \\
Static KD $-$ matched CE, looped & $-0.000994$ & $[-0.001255,\,-0.000713]$ & $-0.000006$ & $[-0.000791,\,+0.000778]$ \\
Staged KD $-$ matched CE, single & $-0.001433$ & $[-0.001555,\,-0.001311]$ & $-0.001009$ & $[-0.001663,\,-0.000293]$ \\
Staged KD $-$ matched CE, looped & $-0.000753$ & $[-0.000905,\,-0.000589]$ & $-0.000456$ & $[-0.001162,\,+0.000302]$ \\
\bottomrule
\end{tabular}

%% file: tables/r43_orders_proxy.tex
\begin{tabular}{lrrr}
\toprule
Comparison & C mean $\pm$ SD & D mean $\pm$ SD & Order interval (95\%) \\
\midrule
Static KD $-$ CE, single & $-0.010778\pm0.000162$ & $-0.010213\pm0.000516$ & $[-0.011073,\,-0.009918]$ \\
Static KD $-$ CE, looped & $+0.001354\pm0.000254$ & $+0.001454\pm0.000368$ & $[+0.001027,\,+0.001781]$ \\
Staged KD $-$ CE, single & $-0.012991\pm0.000440$ & $-0.012517\pm0.000593$ & $[-0.013366,\,-0.012142]$ \\
Staged KD $-$ CE, looped & $+0.000537\pm0.000271$ & $+0.000554\pm0.000070$ & $[+0.000234,\,+0.000857]$ \\
CE $-$ ordinary suffix, single & $+0.028809\pm0.001039$ & $+0.027372\pm0.000813$ & $[+0.027009,\,+0.029173]$ \\
CE $-$ ordinary suffix, looped & $-0.208401\pm0.006421$ & $-0.218805\pm0.011257$ & $[-0.225142,\,-0.202064]$ \\
Joint CE 5\% $-$ Ordinary, loop/single & $-0.004990\pm0.000897$ & $-0.007115\pm0.001732$ & $[-0.007844,\,-0.004260]$ \\
Loop-only $-$ Ordinary, loop/single & $-0.006660\pm0.001342$ & $-0.011436\pm0.000418$ & $[-0.010549,\,-0.007547]$ \\
Static KD $-$ matched CE, single & $-0.000642\pm0.000423$ & $-0.000655\pm0.000393$ & $[-0.001113,\,-0.000185]$ \\
Static KD $-$ matched CE, looped & $-0.001046\pm0.000182$ & $-0.000941\pm0.000316$ & $[-0.001317,\,-0.000670]$ \\
Staged KD $-$ matched CE, single & $-0.001501\pm0.000484$ & $-0.001366\pm0.000597$ & $[-0.002060,\,-0.000807]$ \\
Staged KD $-$ matched CE, looped & $-0.000861\pm0.000387$ & $-0.000645\pm0.000628$ & $[-0.001395,\,-0.000111]$ \\
\bottomrule
\end{tabular}

%% file: tables/r43_effects_wikitext.tex
\begin{tabular}{lrrrr}
\toprule
Comparison & Block mean & Text interval & Time mean & Text interval \\
\midrule
Static KD $-$ CE, single & $-0.006274$ & $[-0.007624,\,-0.004926]$ & $+0.011566$ & $[+0.007802,\,+0.015226]$ \\
Static KD $-$ CE, looped & $+0.011492$ & $[+0.010020,\,+0.012922]$ & $+0.027138$ & $[+0.023097,\,+0.031089]$ \\
Staged KD $-$ CE, single & $-0.012736$ & $[-0.013829,\,-0.011727]$ & $-0.013408$ & $[-0.016308,\,-0.010636]$ \\
Staged KD $-$ CE, looped & $+0.000514$ & $[-0.000877,\,+0.001668]$ & $+0.000249$ & $[-0.003560,\,+0.003763]$ \\
CE $-$ ordinary suffix, single & $+0.026507$ & $[+0.023868,\,+0.029364]$ & $+0.029739$ & $[+0.027105,\,+0.032528]$ \\
CE $-$ ordinary suffix, looped & $-0.325245$ & $[-0.333092,\,-0.316099]$ & $-0.338551$ & $[-0.346352,\,-0.330073]$ \\
Joint CE 5\% $-$ Ordinary, loop/single & $+0.008920$ & $[+0.004362,\,+0.013669]$ & $-0.002228$ & $[-0.007206,\,+0.002519]$ \\
Loop-only $-$ Ordinary, loop/single & $+0.001155$ & $[-0.004077,\,+0.006630]$ & $+0.017321$ & $[+0.011709,\,+0.022939]$ \\
Static KD $-$ matched CE, single & $-0.003633$ & $[-0.004637,\,-0.002640]$ & $+0.004558$ & $[+0.001778,\,+0.007739]$ \\
Static KD $-$ matched CE, looped & $-0.006091$ & $[-0.007358,\,-0.004869]$ & $-0.000593$ & $[-0.003867,\,+0.003075]$ \\
Staged KD $-$ matched CE, single & $-0.000986$ & $[-0.001641,\,-0.000369]$ & $+0.001318$ & $[-0.001163,\,+0.003976]$ \\
Staged KD $-$ matched CE, looped & $-0.001198$ & $[-0.001840,\,-0.000607]$ & $-0.002048$ & $[-0.005051,\,+0.000950]$ \\
\bottomrule
\end{tabular}

%% file: tables/r43_orders_time.tex
\begin{tabular}{lrrr}
\toprule
Comparison & C mean $\pm$ SD & D mean $\pm$ SD & Order interval (95\%) \\
\midrule
Static KD $-$ CE, single & $-0.010861\pm0.000601$ & $-0.010547\pm0.001709$ & $[-0.012581,\,-0.008827]$ \\
Static KD $-$ CE, looped & $+0.002451\pm0.000930$ & $+0.001880\pm0.001388$ & $[+0.000746,\,+0.003585]$ \\
Staged KD $-$ CE, single & $-0.012895\pm0.000679$ & $-0.010554\pm0.001606$ & $[-0.013434,\,-0.010014]$ \\
Staged KD $-$ CE, looped & $+0.000571\pm0.000578$ & $+0.001476\pm0.003136$ & $[-0.002706,\,+0.004753]$ \\
CE $-$ ordinary suffix, single & $+0.029050\pm0.001833$ & $+0.027043\pm0.001572$ & $[+0.026093,\,+0.030000]$ \\
CE $-$ ordinary suffix, looped & $-0.200435\pm0.008434$ & $-0.220287\pm0.010471$ & $[-0.221333,\,-0.199389]$ \\
Joint CE 5\% $-$ Ordinary, loop/single & $-0.006036\pm0.000998$ & $-0.005778\pm0.000526$ & $[-0.006937,\,-0.004876]$ \\
Loop-only $-$ Ordinary, loop/single & $-0.010281\pm0.001490$ & $-0.011478\pm0.001620$ & $[-0.012649,\,-0.009111]$ \\
Static KD $-$ matched CE, single & $-0.000596\pm0.001031$ & $-0.001454\pm0.000988$ & $[-0.002170,\,+0.000120]$ \\
Static KD $-$ matched CE, looped & $+0.000423\pm0.000715$ & $-0.000435\pm0.000504$ & $[-0.000740,\,+0.000728]$ \\
Staged KD $-$ matched CE, single & $-0.001166\pm0.001069$ & $-0.000852\pm0.002583$ & $[-0.003769,\,+0.001751]$ \\
Staged KD $-$ matched CE, looped & $-0.000373\pm0.001090$ & $-0.000540\pm0.002193$ & $[-0.002737,\,+0.001824]$ \\
\bottomrule
\end{tabular}

%% file: tables/presentation_v2/r43_endpoints_confirm_time.tex
\begin{tabular}{lrrr}
\toprule
Method & Updates & Single & Looped \\
\midrule
Ordinary & 4404--4484 & 3.639523 & -- \\
Loop-only & 3417--3461 & 3.807901 & 3.628643 \\
Joint CE 5\% & 2622--2658 & 3.673922 & 3.633616 \\
Joint CE & 2633--2655 & 3.661637 & 3.635265 \\
Online KD & 2169--2188 & 3.650933 & 3.637430 \\
Matched CE & 2502--2520 & 3.651958 & 3.637436 \\
Loop$\to$CE & 3096--3131 & 3.667659 & 3.631022 \\
Loop$\to$KD & 2918--2945 & 3.655935 & 3.632046 \\
Loop$\to$matched CE & 3056--3080 & 3.656944 & 3.632502 \\
Loop$\to$ordinary & 3841--3877 & 3.639613 & 3.841383 \\
\bottomrule
\end{tabular}

%% file: tables/r43_wiki_successes.tex
\begin{tabular}{llllrrrr}
\toprule
Role & Budget & Order & Method & $D_{\rm start}$ & $x$ & $y$ & $d$ \\
\midrule
Fixed & Block & C2 & Matched CE & $-0.041201$ & $-0.002845$ & $-0.016076$ & $-0.013232$ \\
Fixed & Block & C2 & Loop$\to$CE & $-0.041696$ & $-0.003339$ & $-0.036249$ & $-0.032910$ \\
Fixed & Block & C2 & Loop$\to$KD & $-0.050431$ & $-0.012075$ & $-0.029137$ & $-0.017063$ \\
Fixed & Block & C2 & Loop$\to$matched CE & $-0.053064$ & $-0.014708$ & $-0.032329$ & $-0.017621$ \\
Fixed & Block & D2 & Joint CE & $-0.026447$ & $-0.002912$ & $-0.031487$ & $-0.028575$ \\
Fixed & Block & D2 & Matched CE & $-0.029374$ & $-0.005839$ & $-0.017509$ & $-0.011669$ \\
Fixed & Time & D2 & Joint CE & $-0.019570$ & $-0.007605$ & $-0.032482$ & $-0.024877$ \\
Fixed & Time & D2 & Loop$\to$KD & $-0.016262$ & $-0.004297$ & $-0.032411$ & $-0.028114$ \\
Diagnostic & Block & C2 & Loop$\to$CE & $-0.041696$ & $-0.003339$ & $-0.036249$ & $-0.032910$ \\
Diagnostic & Block & C2 & Loop$\to$KD & $-0.050431$ & $-0.012075$ & $-0.029137$ & $-0.017063$ \\
Diagnostic & Block & C2 & Loop$\to$matched CE & $-0.053064$ & $-0.014708$ & $-0.032329$ & $-0.017621$ \\
Diagnostic & Block & D2 & Matched CE & $-0.029374$ & $-0.005839$ & $-0.017509$ & $-0.011669$ \\
Diagnostic & Time & D2 & Joint CE & $-0.019570$ & $-0.007605$ & $-0.032482$ & $-0.024877$ \\
Diagnostic & Time & D2 & Matched CE & $-0.018410$ & $-0.006445$ & $-0.016419$ & $-0.009974$ \\
Diagnostic & Time & D2 & Loop$\to$KD & $-0.016262$ & $-0.004297$ & $-0.032411$ & $-0.028114$ \\
\bottomrule
\end{tabular}

%% file: tables/presentation_v2/r43_criterion_counts.tex
\begin{tabular}{lrrrr}
\toprule
Method & $D_{\rm start}$ & $x$ & $y$ & $d$ \\
\midrule
Ordinary & 6/0/0 & 6/0/0 & 0/6/0 & 0/6/0 \\
Loop-only & 0/6/0 & 0/6/0 & 6/0/0 & 6/0/0 \\
Joint CE 5\% & 0/6/0 & 0/6/0 & 6/0/0 & 6/0/0 \\
Joint CE & 0/6/0 & 0/6/0 & 6/0/0 & 6/0/0 \\
Online KD & 5/0/1 & 0/6/0 & 6/0/0 & 6/0/0 \\
Matched CE & 5/0/1 & 0/6/0 & 4/0/2 & 6/0/0 \\
Loop$\to$CE & 0/6/0 & 0/6/0 & 6/0/0 & 6/0/0 \\
Loop$\to$KD & 0/2/4 & 0/6/0 & 6/0/0 & 6/0/0 \\
Loop$\to$matched CE & 0/4/2 & 0/6/0 & 6/0/0 & 6/0/0 \\
Loop$\to$ordinary & 6/0/0 & 6/0/0 & 0/6/0 & 0/6/0 \\
\bottomrule
\end{tabular}

%% file: tables/r43_selection_counts.tex
\begin{tabular}{llrr}
\toprule
Text & Budget & Fixed & Diagnostic \\
\midrule
Confirmation & Block & 0/60 & 0/60 \\
Confirmation & Time & 0/60 & 0/60 \\
WikiText & Block & 6/60 & 4/60 \\
WikiText & Time & 2/60 & 3/60 \\
\bottomrule
\end{tabular}

%% file: tables/presentation_v2/r43_tasks_primary_accuracy_both.tex
\begin{tabular}{lrrrrrr}
\toprule
& \multicolumn{2}{c}{ARC-Easy} & \multicolumn{2}{c}{PIQA} & \multicolumn{2}{c}{HellaSwag} \\
Method & Single & Looped & Single & Looped & Single & Looped \\
\midrule
Ordinary & 42.88 / 42.97 & 39.97 / 40.41 & 58.41 / 58.66 & 56.96 / 57.68 & 27.18 / 27.24 & 27.13 / 26.89 \\
Loop-only & 41.76 / 41.74 & 43.16 / 43.09 & 57.90 / 58.57 & 57.98 / 58.51 & 27.28 / 27.35 & 27.16 / 27.18 \\
Joint CE 5\% & 43.19 / 43.89 & 43.32 / 43.55 & 58.49 / 58.29 & 58.78 / 58.41 & 27.23 / 27.32 & 27.24 / 27.35 \\
Joint CE & 43.36 / 43.78 & 43.18 / 43.78 & 58.46 / 58.34 & 58.56 / 58.61 & 27.36 / 27.25 & 27.29 / 27.32 \\
Online KD & 43.11 / 43.85 & 43.06 / 43.54 & 58.35 / 58.20 & 58.36 / 58.08 & 27.30 / 27.24 & 27.25 / 27.26 \\
Matched CE & 43.18 / 43.86 & 43.27 / 43.60 & 58.53 / 58.77 & 58.65 / 58.54 & 27.43 / 27.27 & 27.29 / 27.24 \\
Loop$\to$CE & 43.64 / 43.39 & 44.00 / 43.52 & 57.95 / 58.25 & 58.28 / 58.32 & 27.11 / 27.12 & 27.19 / 27.14 \\
Loop$\to$KD & 44.21 / 43.45 & 44.22 / 43.45 & 58.32 / 58.95 & 58.11 / 58.80 & 27.14 / 27.17 & 27.12 / 27.26 \\
Loop$\to$matched CE & 44.21 / 43.15 & 44.36 / 43.55 & 58.18 / 58.41 & 58.30 / 58.70 & 27.21 / 27.01 & 27.21 / 27.19 \\
Loop$\to$ordinary & 43.81 / 43.69 & 41.27 / 41.14 & 58.51 / 58.50 & 57.72 / 57.40 & 27.33 / 27.17 & 27.01 / 27.06 \\
\bottomrule
\end{tabular}

%% file: tables/presentation_v2/r43_tasks_E_common_S0_both.tex
\begin{tabular}{lrrrrrr}
\toprule
& \multicolumn{2}{c}{ARC-Easy} & \multicolumn{2}{c}{PIQA} & \multicolumn{2}{c}{HellaSwag} \\
Method & Single & Looped & Single & Looped & Single & Looped \\
\midrule
Ordinary & 0.00 / 0.00 & 7.15 / 5.90 & 0.00 / 0.00 & 3.28 / 2.72 & 0.00 / 0.00 & 4.45 / 5.13 \\
Loop-only & 2.52 / 2.67 & -0.89 / -0.93 & 1.35 / 1.33 & 0.43 / 0.36 & 1.94 / 2.06 & 0.43 / 0.22 \\
Joint CE 5\% & -1.06 / -1.99 & -1.25 / -1.76 & -0.12 / 0.38 & -0.38 / 0.26 & 0.24 / 0.03 & 0.09 / 0.02 \\
Joint CE & -1.24 / -2.16 & -1.11 / -1.84 & -0.04 / 0.26 & -0.18 / 0.06 & -0.11 / -0.05 & 0.05 / -0.16 \\
Online KD & -1.50 / -2.48 & -1.19 / -2.22 & 0.10 / 0.27 & 0.30 / 0.51 & -0.18 / -0.01 & 0.27 / 0.22 \\
Matched CE & -1.51 / -1.91 & -1.36 / -1.72 & -0.13 / -0.34 & -0.12 / -0.02 & -0.33 / -0.34 & 0.03 / -0.08 \\
Loop$\to$CE & -0.98 / -1.08 & -1.81 / -1.57 & 0.79 / 0.33 & 0.30 / 0.02 & 0.12 / 0.07 & 0.17 / 0.26 \\
Loop$\to$KD & -1.70 / -1.56 & -1.79 / -1.41 & -0.02 / -0.35 & 0.34 / 0.00 & 0.12 / 0.21 & 0.32 / 0.14 \\
Loop$\to$matched CE & -1.88 / -1.00 & -2.20 / -1.25 & 0.35 / 0.01 & 0.24 / -0.04 & -0.06 / 0.37 & 0.17 / 0.38 \\
Loop$\to$ordinary & -1.96 / -2.06 & 4.28 / 4.10 & -0.10 / 0.02 & 2.29 / 2.77 & -0.18 / 0.17 & 3.89 / 4.16 \\
\bottomrule
\end{tabular}

%% file: tables/presentation_v2/r43_diagnostics.tex
\begin{tabular}{lrrrrr}
\toprule
Method & Single NLL & Looped NLL & Forward KL & Fix (\%) & Break (\%) \\
\midrule
Ordinary & 3.6259 & 3.8698 & 0.2111 & 2.6402 & 4.6581 \\
Loop-only & 3.7974 & 3.6163 & 0.1809 & 4.2093 & 2.4029 \\
Joint CE 5\% & 3.6603 & 3.6182 & 0.0489 & 2.0884 & 1.6266 \\
Joint CE & 3.6467 & 3.6189 & 0.0366 & 1.7648 & 1.4287 \\
Online KD & 3.6367 & 3.6207 & 0.0263 & 1.4517 & 1.2664 \\
Matched CE & 3.6379 & 3.6226 & 0.0264 & 1.4531 & 1.2704 \\
Loop$\to$CE & 3.6549 & 3.6179 & 0.0457 & 2.0077 & 1.5863 \\
Loop$\to$KD & 3.6422 & 3.6186 & 0.0351 & 1.6916 & 1.4470 \\
Loop$\to$matched CE & 3.6438 & 3.6193 & 0.0358 & 1.7230 & 1.4521 \\
Loop$\to$ordinary & 3.6268 & 3.8307 & 0.1835 & 2.5616 & 4.1918 \\
\bottomrule
\end{tabular}

%% file: tables/presentation_v2/r43_timing.tex
\begin{tabular}{lrrr}
\toprule
Method & Seconds/update & Allocated GiB & Reserved GiB \\
\midrule
Ordinary & 0.50234 & 5.76 & 6.41 \\
Loop-only & 0.64863 & 6.52 & 7.19 \\
Joint CE 5\% & 0.84690 & 8.51 & 8.81 \\
Joint CE & 0.84702 & 8.51 & 8.81 \\
Online KD & 1.02425 & 8.90 & 9.58 \\
Matched CE & 0.88865 & 8.51 & 8.81 \\
\bottomrule
\end{tabular}

%% file: tables/complete_report_20261011/q4.tex
\begin{tabular}{lrr}
\toprule
Implementation & Matched / Joint CE & Conditional 95\% \\
\midrule
Common weighted-CE entry & 1.000229 & $[0.998948, 1.001511]$ \\
Historical separate entries & 1.051945 & $[1.050699, 1.053194]$ \\
\bottomrule
\end{tabular}

%% file: tables/r43v2_effects_exposure.tex
\begin{tabular}{p{2.8cm}rrrrrr}
\toprule
& \multicolumn{3}{c}{Single-pass output} & \multicolumn{3}{c}{Looped output} \\
Comparison & Mean & Text interval & Order interval & Mean & Text interval & Order interval \\
\midrule
KD $-$ flat & $-453.124$ & $[-456.717,-449.147]$ & $[-498.394,-407.854]$ & $-5.319$ & $[-5.897,-4.783]$ & $[-6.750,-3.888]$ \\
KD $-$ permuted online & $-489.305$ & $[-492.476,-485.731]$ & $[-523.462,-455.149]$ & $-8.159$ & $[-8.638,-7.699]$ & $[-8.678,-7.639]$ \\
Frozen looped $-$ permuted & $-463.495$ & $[-466.594,-460.052]$ & $[-504.144,-422.846]$ & $-10.262$ & $[-10.777,-9.785]$ & $[-10.711,-9.813]$ \\
\midrule
Frozen looped $-$ matched CE & $+28.096$ & $[+27.244,+28.968]$ & $[+22.671,+33.520]$ & $-10.111$ & $[-10.448,-9.786]$ & $[-11.314,-8.908]$ \\
Frozen single-pass $-$ matched CE & $+0.627$ & $[+0.163,+1.106]$ & $[-0.367,+1.622]$ & $-13.449$ & $[-13.789,-13.088]$ & $[-14.204,-12.693]$ \\
KD $-$ frozen looped & $-28.810$ & $[-29.628,-27.996]$ & $[-34.339,-23.281]$ & $+9.064$ & $[+8.676,+9.457]$ & $[+8.297,+9.832]$ \\
KD $-$ frozen single-pass & $-1.342$ & $[-1.734,-0.929]$ & $[-2.563,-0.121]$ & $+12.402$ & $[+12.112,+12.703]$ & $[+11.962,+12.842]$ \\
\midrule
Early-only $-$ matched CE & $+0.019$ & $[-0.257,+0.274]$ & $[-0.241,+0.278]$ & $-0.161$ & $[-0.440,+0.131]$ & $[-0.603,+0.280]$ \\
KD $-$ delayed & $+0.399$ & $[+0.148,+0.654]$ & $[+0.173,+0.625]$ & $+0.147$ & $[-0.145,+0.432]$ & $[-0.340,+0.634]$ \\
Delayed $-$ matched CE & $-1.113$ & $[-1.332,-0.909]$ & $[-1.759,-0.467]$ & $-1.194$ & $[-1.405,-0.981]$ & $[-1.876,-0.511]$ \\
KD $-$ early-only & $-0.733$ & $[-0.992,-0.472]$ & $[-1.464,-0.002]$ & $-0.885$ & $[-1.171,-0.600]$ & $[-1.530,-0.240]$ \\
Early--late interaction & $+0.380$ & $[+0.037,+0.748]$ & $[+0.119,+0.642]$ & $+0.308$ & $[-0.047,+0.683]$ & $[-0.485,+1.102]$ \\
\midrule
KD $-$ matched CE & $-0.714$ & $[-0.955,-0.470]$ & $[-1.406,-0.023]$ & $-1.047$ & $[-1.338,-0.750]$ & $[-1.417,-0.677]$ \\
\bottomrule
\end{tabular}

%% file: tables/r43v2_effects_proxy.tex
\begin{tabular}{p{2.8cm}rrrrrr}
\toprule
& \multicolumn{3}{c}{Single-pass output} & \multicolumn{3}{c}{Looped output} \\
Comparison & Mean & Text interval & Order interval & Mean & Text interval & Order interval \\
\midrule
KD $-$ flat & $-453.124$ & $[-456.717,-449.147]$ & $[-498.394,-407.854]$ & $-5.319$ & $[-5.897,-4.783]$ & $[-6.750,-3.888]$ \\
KD $-$ permuted online & $-489.305$ & $[-492.476,-485.731]$ & $[-523.462,-455.149]$ & $-8.159$ & $[-8.638,-7.699]$ & $[-8.678,-7.639]$ \\
Frozen looped $-$ permuted & $-484.749$ & $[-487.809,-481.232]$ & $[-539.010,-430.487]$ & $-9.417$ & $[-9.879,-8.977]$ & $[-10.189,-8.644]$ \\
\midrule
Frozen looped $-$ matched CE & $+30.733$ & $[+29.266,+32.212]$ & $[+25.920,+35.546]$ & $-3.204$ & $[-4.322,-2.025]$ & $[-5.012,-1.396]$ \\
Frozen single-pass $-$ matched CE & $+1.862$ & $[+0.797,+3.006]$ & $[-0.160,+3.884]$ & $-7.424$ & $[-8.541,-6.192]$ & $[-9.454,-5.393]$ \\
KD $-$ frozen looped & $-31.447$ & $[-32.869,-30.045]$ & $[-35.782,-27.112]$ & $+2.158$ & $[+1.050,+3.230]$ & $[+0.698,+3.617]$ \\
KD $-$ frozen single-pass & $-2.577$ & $[-3.650,-1.582]$ & $[-4.275,-0.878]$ & $+6.377$ & $[+5.235,+7.404]$ & $[+4.666,+8.087]$ \\
\midrule
Early-only $-$ matched CE & $+0.019$ & $[-0.257,+0.274]$ & $[-0.241,+0.278]$ & $-0.161$ & $[-0.440,+0.131]$ & $[-0.603,+0.280]$ \\
KD $-$ delayed & $+0.399$ & $[+0.148,+0.654]$ & $[+0.173,+0.625]$ & $+0.147$ & $[-0.145,+0.432]$ & $[-0.340,+0.634]$ \\
Delayed $-$ matched CE & $-1.113$ & $[-1.332,-0.909]$ & $[-1.759,-0.467]$ & $-1.194$ & $[-1.405,-0.981]$ & $[-1.876,-0.511]$ \\
KD $-$ early-only & $-0.733$ & $[-0.992,-0.472]$ & $[-1.464,-0.002]$ & $-0.885$ & $[-1.171,-0.600]$ & $[-1.530,-0.240]$ \\
Early--late interaction & $+0.380$ & $[+0.037,+0.748]$ & $[+0.119,+0.642]$ & $+0.308$ & $[-0.047,+0.683]$ & $[-0.485,+1.102]$ \\
\midrule
KD $-$ matched CE & $-0.714$ & $[-0.955,-0.470]$ & $[-1.406,-0.023]$ & $-1.047$ & $[-1.338,-0.750]$ & $[-1.417,-0.677]$ \\
\bottomrule
\end{tabular}

%% file: tables/r43v2_effects_time.tex
\begin{tabular}{p{2.8cm}rrrrrr}
\toprule
& \multicolumn{3}{c}{Single-pass output} & \multicolumn{3}{c}{Looped output} \\
Comparison & Mean & Text interval & Order interval & Mean & Text interval & Order interval \\
\midrule
KD $-$ flat & $-479.619$ & $[-483.104,-475.730]$ & $[-527.769,-431.470]$ & $-5.479$ & $[-6.590,-4.530]$ & $[-6.563,-4.394]$ \\
KD $-$ permuted online & $-498.099$ & $[-501.262,-494.650]$ & $[-538.250,-457.947]$ & $-9.290$ & $[-10.254,-8.415]$ & $[-10.383,-8.197]$ \\
Frozen looped $-$ permuted & $-499.475$ & $[-502.889,-495.529]$ & $[-542.431,-456.519]$ & $-10.347$ & $[-11.328,-9.385]$ & $[-12.906,-7.788]$ \\
\midrule
Frozen looped $-$ matched CE & $+27.279$ & $[+25.876,+28.644]$ & $[+25.548,+29.010]$ & $-6.196$ & $[-7.229,-5.244]$ & $[-7.662,-4.729]$ \\
Frozen single-pass $-$ matched CE & $+0.537$ & $[-0.634,+1.595]$ & $[-0.704,+1.778]$ & $-9.884$ & $[-10.908,-8.967]$ & $[-10.866,-8.903]$ \\
KD $-$ frozen looped & $-28.297$ & $[-29.523,-27.089]$ & $[-31.662,-24.933]$ & $+6.274$ & $[+5.395,+7.101]$ & $[+4.986,+7.562]$ \\
KD $-$ frozen single-pass & $-1.555$ & $[-2.475,-0.653]$ & $[-3.003,-0.108]$ & $+9.962$ & $[+9.077,+10.820]$ & $[+8.035,+11.890]$ \\
\midrule
Early-only $-$ matched CE & $+0.250$ & $[-0.360,+0.841]$ & $[-1.089,+1.588]$ & $+0.102$ & $[-0.508,+0.704]$ & $[-1.253,+1.458]$ \\
KD $-$ delayed & $-0.651$ & $[-1.318,+0.038]$ & $[-2.684,+1.382]$ & $-0.987$ & $[-1.676,-0.280]$ & $[-2.646,+0.671]$ \\
Delayed $-$ matched CE & $-0.367$ & $[-0.991,+0.266]$ & $[-2.142,+1.408]$ & $+1.066$ & $[+0.446,+1.724]$ & $[-1.125,+3.256]$ \\
KD $-$ early-only & $-1.268$ & $[-1.974,-0.594]$ & $[-2.601,+0.065]$ & $-0.024$ & $[-0.710,+0.637]$ & $[-1.611,+1.562]$ \\
Early--late interaction & $-0.901$ & $[-1.742,-0.079]$ & $[-1.980,+0.179]$ & $-1.090$ & $[-1.952,-0.276]$ & $[-3.171,+0.991]$ \\
\midrule
KD $-$ matched CE & $-1.018$ & $[-1.646,-0.389]$ & $[-3.044,+1.008]$ & $+0.078$ & $[-0.527,+0.733]$ & $[-1.700,+1.856]$ \\
\bottomrule
\end{tabular}

%% file: tables/r43v2_retention_time.tex
\begin{tabular}{lrrrr}
\toprule
Method & $D_{\rm start}$ & $x$ & $y$ & $d$ \\
\midrule
Online KD & $+0.002040$ & $+0.011216$ & $-0.001728$ & $-0.012945$ \\
Matched CE & $+0.003058$ & $+0.012235$ & $-0.001806$ & $-0.014041$ \\
Frozen looped & $+0.030337$ & $+0.039514$ & $-0.008002$ & $-0.047515$ \\
Frozen single-pass & $+0.003595$ & $+0.012772$ & $-0.011691$ & $-0.024462$ \\
\bottomrule
\end{tabular}

%% file: tables/r46_ppl_scales.tex
\begin{tabular}{lrrrrrr}
\toprule
NLL increase & 0.0005 & 0.001 & 0.002 & 0.005 & 0.010 & 0.020 \\
\midrule
PPL increase (\%) & 0.0500 & 0.1001 & 0.2002 & 0.5013 & 1.0050 & 2.0201 \\
\bottomrule
\end{tabular}

%% file: tables/r46_retention_scales.tex
\begin{tabular}{llrrrrrr}
\toprule
Study / reference & Budget & $0$ & $.001$ & $.002$ & $.005$ & $.01$ & $.02$ \\
\midrule
Supervision / fixed & Block & 0/60/0 & 0/60/0 & 0/60/0 & 0/60/0 & 0/57/3 & 17/34/9 \\
Supervision / fixed & Time & 0/60/0 & 0/60/0 & 0/60/0 & 0/60/0 & 1/55/4 & 18/35/7 \\
Target / fixed & Block & 0/102/0 & 0/102/0 & 0/102/0 & 0/102/0 & 0/91/11 & 35/58/9 \\
Target / fixed & Time & 0/102/0 & 0/102/0 & 0/102/0 & 0/102/0 & 0/89/13 & 28/56/18 \\
Target / selected & Block & 0/102/0 & 0/102/0 & 0/102/0 & 0/102/0 & 0/91/11 & 35/58/9 \\
Target / selected & Time & 0/102/0 & 0/102/0 & 0/102/0 & 0/102/0 & 0/89/13 & 28/56/18 \\
\bottomrule
\end{tabular}

%% file: tables/r46_equivalence_scales.tex
\begin{tabular}{lllrrrr}
\toprule
Budget & Output & Interval & $.0005$ & $.001$ & $.002$ & $.005$ \\
\midrule
Equal updates & Single & Text family & U & U & S & S \\
Equal updates & Single & Order 95\% & U & U & S & S \\
Block & Single & Text family & V & U & U & S \\
Block & Single & Order 95\% & U & U & U & S \\
Time & Single & Text family & U & U & S & S \\
Time & Single & Order 95\% & U & U & S & S \\
\bottomrule
\end{tabular}

%% file: tables/presentation_v2/minimal_teacher_time.tex
\begin{tabular}{lrrrr}
\toprule
& \multicolumn{2}{c}{Single-pass output} & \multicolumn{2}{c}{Looped output} \\
Teacher / method & $\Delta$NLL & Order 95\% interval & $\Delta$NLL & Order 95\% interval \\
\midrule
Online KD & $-1.018$ & $[-3.044,+1.008]$ & $+0.078$ & $[-1.700,+1.856]$ \\
Frozen initial looped output & $+27.279$ & $[+25.548,+29.010]$ & $-6.196$ & $[-7.662,-4.729]$ \\
\textbf{Frozen-original KD (ours)} & $+0.537$ & $[-0.704,+1.778]$ & $\boldsymbol{-9.884}$ & $\boldsymbol{[-10.866,-8.903]}$ \\
\bottomrule
\end{tabular}

%% file: tables/new_results_20261011/selection.tex
\begin{tabular}{lrrr}
\toprule
Tolerance $\tau$ & CE & Online KD & Frozen-original KD \\
\midrule
0.001 & None & None & 0.4 \\
0.005 & 0.8 & 0.8 & 0.4 \\
0.01 & 0.4 & 0.4 & 0.4 \\
0.02 & 0.4 & 0.2 & 0.4 \\
\bottomrule
\end{tabular}

%% file: tables/new_results_20261011/full_grid.tex
\begin{tabular}{lrrrrrc}
\toprule
Family & $m$ & $D_{\rm start}$ & $x$ & $y$ & $d$ & All four \\
\midrule
Label-only CE & 0.2 & $+0.003687$ & $+0.011903$ & $-0.002327$ & $-0.014230$ & 0/6 \\
 & 0.4 & $-0.002705$ & $+0.005511$ & $+0.000808$ & $-0.004703$ & 2/6 \\
 & 0.8 & $-0.006054$ & $+0.002162$ & $+0.003877$ & $+0.001715$ & 0/6 \\
 & 1.6 & $-0.006226$ & $+0.001990$ & $+0.008381$ & $+0.006392$ & 0/6 \\
\midrule
Online KD & 0.2 & $+0.002618$ & $+0.010833$ & $-0.002341$ & $-0.013174$ & 0/6 \\
 & 0.4 & $-0.003185$ & $+0.005030$ & $+0.001023$ & $-0.004008$ & 1/6 \\
 & 0.8 & $-0.005858$ & $+0.002357$ & $+0.004443$ & $+0.002086$ & 0/6 \\
 & 1.6 & $-0.006435$ & $+0.001781$ & $+0.008654$ & $+0.006874$ & 0/6 \\
\midrule
Frozen-original KD (ours) & 0.2 & $+0.004206$ & $+0.012421$ & $-0.012490$ & $-0.024911$ & 0/6 \\
 & 0.4 & $-0.008741$ & $-0.000526$ & $-0.008398$ & $-0.007872$ & 6/6 \\
 & 0.8 & $-0.011328$ & $-0.003113$ & $-0.002193$ & $+0.000919$ & 0/6 \\
 & 1.6 & $-0.010265$ & $-0.002049$ & $+0.004300$ & $+0.006349$ & 0/6 \\
\bottomrule
\end{tabular}

%% file: tables/complete_report_20261011/grid_comparisons.tex
\begin{tabular}{llrr}
\toprule
CE weight $m$ & Output & Frozen $m=0.4$ $-$ CE & Text 99.9621\% \\
\midrule
0.2 & Single & $-12.428$ & $[-13.738, -11.187]$ \\
 & Loop & $-6.071$ & $[-7.523, -4.772]$ \\
0.4 & Single & $-6.036$ & $[-7.604, -4.616]$ \\
 & Loop & $-9.206$ & $[-10.846, -7.817]$ \\
0.8 & Single & $-2.687$ & $[-3.917, -1.376]$ \\
 & Loop & $-12.274$ & $[-13.609, -10.994]$ \\
1.6 & Single & $-2.515$ & $[-3.530, -1.512]$ \\
 & Loop & $-16.779$ & $[-17.828, -15.781]$ \\
\bottomrule
\end{tabular}

%% file: tables/new_results_20261011/selected_intervals.tex
\begin{tabular}{llrrr}
\toprule
Recipe & Criterion & Mean & Text 99.5833\% & Order 95\% \\
\midrule
Label-only CE & $D_{\rm start}$ & $-0.006291$ & $[-0.007798, -0.004773]$ & $[-0.010153, -0.002429]$ \\
 & $x$ & $+0.001929$ & $[+0.001013, +0.002827]$ & $[-0.001274, +0.005132]$ \\
 & $y$ & $+0.003571$ & $[+0.002640, +0.004477]$ & $[+0.000503, +0.006639]$ \\
 & $d$ & $+0.001642$ & $[+0.001281, +0.002090]$ & $[+0.000980, +0.002305]$ \\
\midrule
Online KD & $D_{\rm start}$ & $-0.005657$ & $[-0.007281, -0.003673]$ & $[-0.017427, +0.006112]$ \\
 & $x$ & $+0.002563$ & $[+0.001586, +0.003588]$ & $[-0.009883, +0.015009]$ \\
 & $y$ & $+0.004641$ & $[+0.003701, +0.005663]$ & $[-0.006969, +0.016251]$ \\
 & $d$ & $+0.002078$ & $[+0.001816, +0.002381]$ & $[+0.001661, +0.002495]$ \\
\midrule
Frozen-original KD (ours) & $D_{\rm start}$ & $-0.008953$ & $[-0.010308, -0.007513]$ & $[-0.012569, -0.005336]$ \\
 & $x$ & $-0.000733$ & $[-0.001940, +0.000446]$ & $[-0.004150, +0.002684]$ \\
 & $y$ & $-0.008327$ & $[-0.009513, -0.007225]$ & $[-0.013522, -0.003133]$ \\
 & $d$ & $-0.007595$ & $[-0.008033, -0.007064]$ & $[-0.009383, -0.005806]$ \\
\bottomrule
\end{tabular}

%% file: tables/new_results_20261011/selected_effects.tex
\begin{tabular}{llrrr}
\toprule
Comparison & Output & Mean & Text 98.75\% & Order 95\% \\
\midrule
Online KD $-$ selected CE & Single & $+0.000634$ & $[-0.000196, +0.001512]$ & $[-0.015180, +0.016448]$ \\
Online KD $-$ selected CE & Looped & $+0.001070$ & $[+0.000273, +0.001867]$ & $[-0.010922, +0.013061]$ \\
Frozen-original KD (ours) $-$ selected CE & Single & $-0.002662$ & $[-0.003764, -0.001667]$ & $[-0.011338, +0.006015]$ \\
Frozen-original KD (ours) $-$ selected CE & Looped & $-0.011899$ & $[-0.013084, -0.010875]$ & $[-0.020206, -0.003592]$ \\
\bottomrule
\end{tabular}

%% file: tables/new_results_20261011/selected_orders.tex
\begin{tabular}{llrrrrc}
\toprule
Recipe & Order & $D_{\rm start}$ & $x$ & $y$ & $d$ & All four \\
\midrule
Label-only CE & C2 & $-0.004975$ & $+0.002416$ & $+0.004632$ & $+0.002216$ & No \\
 & C3 & $-0.006405$ & $+0.000781$ & $+0.002625$ & $+0.001844$ & No \\
 & D2 & $-0.005671$ & $+0.003361$ & $+0.004528$ & $+0.001166$ & No \\
 & D3 & $-0.008114$ & $+0.001157$ & $+0.002501$ & $+0.001344$ & No \\
\midrule
Online KD & C2 & $-0.005499$ & $+0.001892$ & $+0.004260$ & $+0.002368$ & No \\
 & C3 & $-0.005456$ & $+0.001731$ & $+0.004315$ & $+0.002585$ & No \\
 & D2 & $-0.007691$ & $+0.001341$ & $+0.003165$ & $+0.001824$ & No \\
 & D3 & $-0.003984$ & $+0.005287$ & $+0.006824$ & $+0.001536$ & No \\
\midrule
Frozen-original KD (ours) & C2 & $-0.009372$ & $-0.001981$ & $-0.008837$ & $-0.006855$ & Yes \\
 & C3 & $-0.010733$ & $-0.003547$ & $-0.011038$ & $-0.007491$ & Yes \\
 & D2 & $-0.008001$ & $+0.001032$ & $-0.007041$ & $-0.008072$ & Yes \\
 & D3 & $-0.007705$ & $+0.001566$ & $-0.006394$ & $-0.007960$ & Yes \\
\bottomrule
\end{tabular}

%% file: tables/new_results_20261011/receivers.tex
\begin{tabular}{llrr}
\toprule
KL receiving path & Matched CE $(\mathcal L_s,\mathcal L_c)$ & Single difference & Looped difference \\
\midrule
Single pass & $(0.2,0.9)$ & $+0.000519$ & $-0.010163$ \\
Looped output & $(0.1,1.0)$ & $-0.013423$ & $-0.009639$ \\
Both outputs & $(0.15,0.95)$ & $-0.005932$ & $-0.008324$ \\
\bottomrule
\end{tabular}

%% file: tables/new_results_20261011/extensions.tex
\begin{tabular}{lrrrrl}
\toprule
Configuration and recipe & $n$ & $x$ & $y$ & Looped $-$ CE & CE reference \\
\midrule
Main $8\to12$, selected $m=0.4$ & 4 & $-0.000733$ & $-0.008327$ & $-0.011899$ & Selected CE \\
Main selected recipe, WikiText & 4 & $+0.016779$ & $+0.008051$ & -- & -- \\
Deeper $8\to16$, fixed coefficients & 4 & $+0.015635$ & $-0.010416$ & $-0.010057$ & Fixed CE \\
Native R1$\to$R4, fixed coefficients & 6 & $+0.006058$ & $+0.003877$ & $-0.002413$ & Fixed CE \\
\bottomrule
\end{tabular}

%% file: tables/complete_report_20261011/receiver_interactions.tex
\begin{tabular}{lllrrr}
\toprule
Receivers A $-$ B & Contrast & Output & Mean & Text 99.5833\% & Order 95\% \\
\midrule
Single $-$ loop receiver & Direct & Single & $+2.802$ & $[+1.953, +3.635]$ & $[+0.871, +4.733]$ \\
 & Direct & Loop & $+1.283$ & $[+0.519, +2.082]$ & $[+0.130, +2.436]$ \\
 & Interaction & Single & $+13.941$ & $[+12.895, +14.892]$ & $[+12.327, +15.556]$ \\
 & Interaction & Loop & $-0.524$ & $[-1.389, +0.398]$ & $[-2.400, +1.352]$ \\
\midrule
Single $-$ both receivers & Direct & Single & $+2.241$ & $[+1.392, +3.093]$ & $[+0.002, +4.479]$ \\
 & Direct & Loop & $-0.632$ & $[-1.529, +0.282]$ & $[-2.432, +1.167]$ \\
 & Interaction & Single & $+6.450$ & $[+5.491, +7.450]$ & $[+4.228, +8.672]$ \\
 & Interaction & Loop & $-1.839$ & $[-2.814, -0.821]$ & $[-3.723, +0.044]$ \\
\midrule
Loop $-$ both receivers & Direct & Single & $-0.561$ & $[-1.274, +0.173]$ & $[-2.716, +1.594]$ \\
 & Direct & Loop & $-1.915$ & $[-2.630, -1.143]$ & $[-4.490, +0.660]$ \\
 & Interaction & Single & $-7.491$ & $[-8.292, -6.713]$ & $[-10.505, -4.477]$ \\
 & Interaction & Loop & $-1.315$ & $[-2.149, -0.533]$ & $[-3.871, +1.240]$ \\
\bottomrule
\end{tabular}

%% file: tables/complete_report_20261011/structure_intervals.tex
\begin{tabular}{llrrr}
\toprule
Fixed-recipe configuration & Output & Frozen $-$ CE & Text 99.1667\% & Order 95\% \\
\midrule
HF $8\to16$ & Single & $+3.189$ & $[+2.120, +4.350]$ & $[-1.514, +7.892]$ \\
 & Loop & $-10.057$ & $[-11.202, -8.869]$ & $[-17.355, -2.759]$ \\
Native Parcae & Single & $-2.250$ & $[-2.719, -1.815]$ & $[-2.495, -2.005]$ \\
 & Loop & $-2.413$ & $[-2.891, -1.959]$ & $[-2.990, -1.836]$ \\
\bottomrule
\end{tabular}

%% file: tables/complete_report_20261011/native_noise.tex
\begin{tabular}{lrrr}
\toprule
Fixed recipe & Noise view & $x$ & $y$ \\
\midrule
Label-only CE & 0 & $+8.307$ & $+6.290$ \\
 & 1 & $+8.316$ & $+6.302$ \\
 & 2 & $+8.306$ & $+6.316$ \\
Online KD & 0 & $+9.084$ & $+7.371$ \\
 & 1 & $+9.083$ & $+7.383$ \\
 & 2 & $+9.080$ & $+7.381$ \\
Frozen-original KD & 0 & $+6.058$ & $+3.877$ \\
 & 1 & $+6.049$ & $+3.892$ \\
 & 2 & $+6.043$ & $+3.879$ \\
\bottomrule
\end{tabular}

%% file: tables/new_results_20261011/tasks.tex
\begin{tabular}{llrrr}
\toprule
Recipe & Output & ARC-Easy & PIQA & HellaSwag \\
\midrule
Ordinary & Single & 43.08 & 58.65 & 27.32 \\
Selected Label-only CE & Single & 44.41 & 58.71 & 27.48 \\
Selected Label-only CE & Looped & 43.70 & 58.60 & 27.42 \\
Selected Online KD & Single & 44.03 & 58.31 & 27.20 \\
Selected Online KD & Looped & 43.82 & 58.30 & 27.22 \\
Selected Frozen-original KD (ours) & Single & 44.14 & 58.72 & 27.06 \\
Selected Frozen-original KD (ours) & Looped & 43.64 & 58.73 & 27.22 \\
\bottomrule
\end{tabular}

%% file: tables/complete_report_20261011/task_retention.tex
\begin{tabular}{lllrrr}
\toprule
Task & Selected recipe & Output & $E$ (pp) & Item 95\% (pp) & Order 95\% (pp) \\
\midrule
ARC-Easy & Label-only CE, $m=0.8$ & Single & $-2.39$ & $[-3.78, -1.03]$ & $[-5.72, +0.94]$ \\
 & Label-only CE, $m=0.8$ & Looped & $-1.67$ & $[-3.03, -0.30]$ & $[-5.05, +1.71]$ \\
 & Online KD, $m=0.8$ & Single & $-2.25$ & $[-3.66, -0.87]$ & $[-9.77, +5.26]$ \\
 & Online KD, $m=0.8$ & Looped & $-1.75$ & $[-3.16, -0.36]$ & $[-13.64, +10.15]$ \\
 & Frozen-original KD, $m=0.4$ & Single & $-3.53$ & $[-4.84, -2.26]$ & $[-18.12, +11.06]$ \\
 & Frozen-original KD, $m=0.4$ & Looped & $-2.83$ & $[-4.19, -1.52]$ & $[-8.93, +3.27]$ \\
\midrule
PIQA & Label-only CE, $m=0.8$ & Single & $-0.09$ & $[-1.02, +0.83]$ & $[-4.79, +4.61]$ \\
 & Label-only CE, $m=0.8$ & Looped & $+0.21$ & $[-0.70, +1.12]$ & $[-2.52, +2.94]$ \\
 & Online KD, $m=0.8$ & Single & $+0.12$ & $[-0.80, +1.06]$ & $[-2.43, +2.67]$ \\
 & Online KD, $m=0.8$ & Looped & $+0.05$ & $[-0.90, +0.98]$ & $[-2.37, +2.46]$ \\
 & Frozen-original KD, $m=0.4$ & Single & $-0.77$ & $[-1.65, +0.12]$ & $[-2.30, +0.77]$ \\
 & Frozen-original KD, $m=0.4$ & Looped & $-0.47$ & $[-1.34, +0.40]$ & $[-1.07, +0.14]$ \\
\midrule
HellaSwag & Label-only CE, $m=0.8$ & Single & $-0.65$ & $[-1.38, +0.07]$ & $[-5.28, +3.98]$ \\
 & Label-only CE, $m=0.8$ & Looped & $-0.31$ & $[-1.04, +0.42]$ & $[-3.27, +2.64]$ \\
 & Online KD, $m=0.8$ & Single & $+0.24$ & $[-0.49, +0.96]$ & $[-0.65, +1.12]$ \\
 & Online KD, $m=0.8$ & Looped & $+0.37$ & $[-0.35, +1.10]$ & $[-0.70, +1.45]$ \\
 & Frozen-original KD, $m=0.4$ & Single & $-0.73$ & $[-1.45, -0.02]$ & $[-3.59, +2.13]$ \\
 & Frozen-original KD, $m=0.4$ & Looped & $-0.34$ & $[-1.06, +0.37]$ & $[-6.53, +5.85]$ \\
\bottomrule
\end{tabular}